\documentclass{article}
\usepackage{iclr2026_conference,times}
\usepackage[hypertexnames=false]{hyperref}
\usepackage{url}
\usepackage{graphicx}
\usepackage{booktabs}
\usepackage{multirow}
\usepackage{amsmath,amssymb}
\usepackage{algorithm}
\usepackage{algpseudocode}
\usepackage[table]{xcolor}
\usepackage{placeins}
\usepackage{float}
\usepackage{wrapfig}
\usepackage{capt-of}
\usepackage{microtype}

\newcommand{\gray}[1]{\cellcolor{gray!20}#1}

\title{Improving Cross-Problem Vehicle Routing with Locally Augmented Preferences and Representation Disentanglement}

\author{%
Arthur Corr\^ea\textsuperscript{*}\thanks{Corresponding author.},~
Paulo Nascimento,~
Samuel Moniz \\
University of Coimbra, CEMMPRE, ARISE \\
Department of Mechanical Engineering, Coimbra 3030-788, Portugal \\
\texttt{\{ajpcorrea, pnascimento, samuel.moniz\}@dem.uc.pt}
}

\iclrfinalcopy

\begin{document}
\maketitle

\begin{abstract}
Multi-task vehicle routing problem (VRP) solvers seek to handle multiple VRP variants within a single unified model, avoiding the need to train a separate model for every variant. In spite of recent progress, current approaches remain limited on two fronts. On the training side, reinforcement learning suffers from reward-scale disparities and shrinking advantage signals as policies improve, whereas preference optimization stagnates once sampled tours become near-identical and thus fundamentally limited by the quality of the policy's own generated solutions, leaving both paradigms with weak supervision as training progresses. On the architecture side, existing fully shared encoders entangle constraint-dependent representations across heterogeneous variants, which limits generalization. We address these gaps with two model-agnostic contributions. First, we propose Preference Optimization with Locally Augmented Refinement (POLAR), a novel training algorithm that applies a local search refinement pass to the best decoded tour before forming preference pairs, yielding much more informative pairwise margins. Second, a Progressive Layered Extraction (PLE) encoder routes each encoder layer through one shared expert and a set of task-specific experts via a gating mechanism, progressively separating common routing structure from constraint-specific encodings. Through extensive experiments on various VRP variants, we show that POLAR and PLE together elevate the current state-of-the-art among neural multi-task solvers. We reduce the average gap to reference solutions by 21.3\% relative to the strongest published baseline on 16 in-distribution variants, and outperform prior neural methods on 27 out of 32 unseen variants. Ablation studies confirm the efficacy of each contribution, showing that both improve cross-problem generalization across multiple backbone model architectures.
\end{abstract}

	\section{Introduction}
	\label{sec:intro}
	
	The Vehicle Routing Problem (VRP) is a critical Combinatorial Optimization (CO) problem in modern logistics and supply-chain decision support \citep{Cattaruza2017,Wang2020logistics}. In essence, it entails finding the optimal set of routes for a fleet of vehicles to serve geographically dispersed customers, with the objective typically being to minimize the total travel distance. Historically, VRPs have been mainly tackled by exact algorithms and meta-heuristics \citep{Laporte1992}. While exact methods guarantee optimality, they scale poorly on instances larger than one hundred nodes \citep{Pessoa2020}. Conversely, meta-heuristics present greater scalability and speed, but their design requires expert tuning. More recently, neural constructive solvers have emerged as a compelling alternative for CO problems \citep{Bengio2021}. By leveraging deep learning to learn highly effective policies, these methods achieve much more reasonable computational times than traditional approaches, without requiring specialized domain knowledge. In practice, however, VRPs are subject to diverse operational constraints, such as vehicle capacity, time windows and backhauls, whose combinations give rise to numerous variants of the problem \citep{Braekers2016}.
	
	Recent work has addressed this heterogeneity through unified Multi-Task Learning (MTL) models that learn shared representations across multiple VRP variants \citep{Liu2024,Zhou2024,Berto2025,Li2025}. This paradigm is attractive from the perspective of decision support systems and practitioners, since a single deployable model can support diverse routing scenarios without retraining a separate solver for every operational configuration. Nevertheless, some limitations still prevent MTL models from achieving better generalization. First, training remains dominated by Reinforcement Learning (RL). Although RL has long been the gold standard training paradigm in neural-based CO, it suffers from some drawbacks: (i) policies become less informative as they improve, since the advantage gap between sampled solutions and their baseline shrinks, which can slow down convergence; (ii) the different reward scales between VRP variants produce uneven gradient magnitudes, which is problematic in the MTL setting since gradients from one task can dominate and suppress updates for others \citep{Hessel2019}. Alternative training paradigms, such as self-labeling from the best sampled tour \citep{Corsini2024} reduce reliance on scalar rewards, but discard most generated trajectories, suffering from limited sample efficiency. Preference Optimization (PO) \citep{Pan2025,Correa2026filmmed} offers a more attractive compromise. By learning from relative rankings, the supervision is inherently insensitive to heterogeneous cost scales. Yet, PO is only as informative as the tours being compared. When the policy generates increasingly similar solutions, pairwise preferences carry small cost margins and learning can stagnate for reasons analogous to the advantage collapse in RL. This observation highlights a critical opportunity: improving the quality of candidate solutions used for preference learning. If policy-generated solutions could be efficiently refined in an \textit{a posteriori} manner (without distorting the underlying policy distribution), the resulting higher-quality pairwise comparisons could provide stronger learning signals. Such an approach has the potential to navigate towards much more promising regions of the solution space by escaping local minima, narrowing the performance gaps between neural methods and traditional optimization algorithms even more.
	
	Second, most existing MTL architectures adopt fully shared representations, wherein a single backbone is trained jointly across all problem variants. This design assumes that a single unified latent space can capture the heterogeneous structures of diverse constraints \citep{Ruder2017}. While this architectural choice enables parameter efficiency, it introduces some limitations. In particular, fully shared models are more prone to the gradient interference challenge \citep{Yu2020}, which can entangle conflicting task signals in shared parameters and yield suboptimal representations across all tasks. Without mechanisms that separate common routing geometry from constraint-specific computation, dominant patterns learned for one variant can suppress or overwrite those required by another. Recent architectural proposals mitigate this issue through soft parameter sharing and feature-wise linear modulation \citep{Wang2026,Correa2026}. Nevertheless, representation entanglement remains a fundamental challenge in cross-problem VRP learning.
	
	%This paper is motivated by a hybrid principle that addresses the aforementioned limitations jointly: classical Local Search (LS) algorithms can strengthen both learning and inference. During training, refining decoded tours before preference comparison increases the margin between preferred and dispreferred solutions, yielding denser and more informative PO supervision. At inference, the same principle motivates a complementary design. If the neural-based model by itself is capable of landing in promising regions of the solution space, then generated tours can anchor a meta-heuristic algorithm instead of relying on a generic constructive heuristic to initialize solutions. Our experiments further show that the resulting performance gap scales almost invariably with the quality of the initial solution. This finding underscores a broader insight for future neural CO research: as neural-based solvers evolve and mature, their solutions are now of sufficient quality to serve as reliable anchors for meta-heuristics, enabling a new class of hybrid algorithms that surpass conventional methods.
	
	Building on the aforementioned gaps, we develop a unified framework comprising POLAR (Preference Optimization with Local Augmented Refinement) for training-time refinement and a Progressive Layered Extraction (PLE) encoder for effective representation disentanglement. Our main contributions are as follows:

	\begin{itemize}
		\item We propose POLAR, a novel training algorithm for MTL models, combining PO with an LS refinement step on the best decoded tour per instance. By widening pairwise preference margins at a modest CPU cost, POLAR significantly improves convergence and generalization over REINFORCE and PO-only training across multiple MTL backbones.
		\item We introduce a PLE encoder motivated by gradient interference and representation entanglement across heterogeneous constraints. A hierarchy of shared and task-specific experts, combined with a learned gating mechanism, progressively separates common routing structure from variant-specific signals without having to train separate models for each constraint type.
		\item Extensive experiments on synthetic and benchmark instances demonstrate that POLAR and PLE elevate the existing state-of-the-art on MTL models. Through comprehensive ablations, we validate the effectiveness and individual impact of our contributions. Moreover, we show that both POLAR and PLE are architecture-agnostic, allowing seamless integration into a wide range of MTL backbones while consistently improving generalization.
	\end{itemize}

	The rest of this paper is organized as follows. Section~\ref{sec_preliminaries} describes the problem setting. Section~\ref{sec_methodology} presents the PLE encoder and the POLAR training algorithm. Section~\ref{sec_experiments} presents the experiments, including ablations. Section~\ref{sec_conclusion} concludes. Related work (Appendix~\ref{app:related}), a detailed architecture overview and further analyses are deferred to the appendix.

\section{Problem description}
	\label{sec_preliminaries}
	
	Following the framework established by \citet{Berto2025}, this work addresses 48 distinct VRP variants. A standard Capacitated VRP (CVRP) instance with $n$ customers is defined on a graph $\mathcal{G} = \{\mathcal{V}, \mathcal{E}\}$. Here, $\mathcal{V} = \{v_0, v_1, ..., v_n\}$ denotes the set of nodes, with $v_0$ being the depot. Each node $v_i \in \mathcal{V}$ is characterized by 2D coordinates $(x_i, y_i)$, and each customer $v_i$, $i>0$, has demand $\delta_i > 0$. The edge set $\mathcal{E} = \{e_{ij}: i, j \in \mathcal{V}, i \neq j\}$ represents all possible connections between these nodes, with each edge $e_{ij}$ having an associated Euclidean travel cost $c_{ij}$. A fleet of vehicles, each with capacity $C$, is stationed at the depot to serve the customers. A solution $\tau$ consists of individual vehicle's routes, with each one departing from the depot, visiting a sequence of customers, and returning to the depot in the end. $\tau$ is feasible as long as all customers are visited exactly once by a single vehicle, and the cumulative demand of any individual route does not exceed $C$. The objective is to find the optimal solution $\tau^*$ that minimizes the total distance traveled by the entire fleet.

	On top of the CVRP, we consider six additional constraints that combine to yield 48 distinct variants: (i) \textit{Backhauls} (\textit{B}): The CVRP involves only linehaul customers, i.e., deliveries that effectively reduce the vehicle's load throughout a route. In the VRP with Backhauls (VRPB), some customers (known as backhaul customers) require pickups instead of deliveries, thereby increasing the vehicle's load. In this setting, a vehicle must service all linehaul customers on its route before visiting any backhaul customers. Additionally, the vehicle load must remain within capacity $C$ at all times; (ii) \textit{Open Routes} (\textit{O}): In the VRP with Open Routes (OVRP) a vehicle is not required to return to the depot after completing its route. As a result, the final depot-return arc is omitted from the route cost and feasibility checks; (iii) \textit{Distance Limits} (\textit{L}): The VRP with Distance Limits (VRPL) imposes a maximum threshold $D$ on the total distance or duration of each route. Consequently, the cumulative cost of the arcs traveled by a vehicle, including the return trip to the depot (when required), must not exceed $D$; (iv) \textit{Time Windows} (\textit{TW}): In the VRP with Time Windows (VRPTW), each customer $v_i$ is assigned a service duration $t^s_i$ and a time window $[t^e_i, t^l_i]$ during which service must commence. If a vehicle arrives before $t^e_i$, it must wait until the time window opens. Additionally, the depot has a fixed closing time $t^l_0$, requiring all vehicles to return before this time; (v) \textit{Mixed Backhauls} (\textit{MB}): Unlike the strict precedence in standard backhauls, in the VRP with Mixed Backhauls (VRPMB), vehicles may visit linehaul and backhaul customers in any order. However, as in the VRPB, the vehicle's current load must never exceeds its maximum capacity $C$ throughout a route; (vi) \textit{Multi-Depots} (\textit{MD}): In the Multi-Depot VRP (MDVRP), instances contain multiple depots instead of a single one. Additionally, each vehicle is required to start and end its route at the same depot.
	
	During decoding, feasibility is enforced at all times through action masks. Following prior work \citep{Berto2025,Li2025,Liu2025}, we train on 16 variants comprising all combinations of open routes, time windows, backhauls, and distance limits. We evaluate these same 16 variants in the main experiments and additionally report zero-shot transfer to 32 unseen variants with mixed backhauls and multi-depots in Appendix~\ref{app:zeroshot}.

	\section{Methodology}
	\label{sec_methodology}
	
	Our framework has two complementary components. The first is architectural: a PLE encoder that decomposes each layer into shared and task-specific experts, while keeping the pathways coupled through gated residual connections. This design is motivated by gradient interference and representation entanglement when multiple VRP variants are trained jointly. The second is algorithmic: POLAR, a preference-based training procedure that refines the best decoded tour with an LS before preference comparison. This design targets the shrinking preference margins that arise once a PO policy begins to converge. In this Section, we present the implementation details of both contributions. Solution construction follows the encoder--decoder Markov decision process recalled in Appendix~\ref{app:learning}. Every expert block uses the self-attention operator reviewed in Appendix~\ref{app:attention}. Details on input featurization, encoder, and the autoregressive decoder are specified in Appendix~\ref{app:architecture}.

	\subsection{Progressive layered extraction encoder}
	\label{sec:model_architecture}
	
	We follow the heavy-encoder, light-decoder design that has become standard in modern multi-task VRP solvers, but replace the conventional fully shared encoder stack with a progressive hierarchy of shared and task experts. Our goal is not merely to increase the encoder's capacity, but to give heterogeneous constraint signals separate computational pathways before decoding begins.
	
	\begin{figure}[t]
		\centering
		\includegraphics[width=\linewidth]{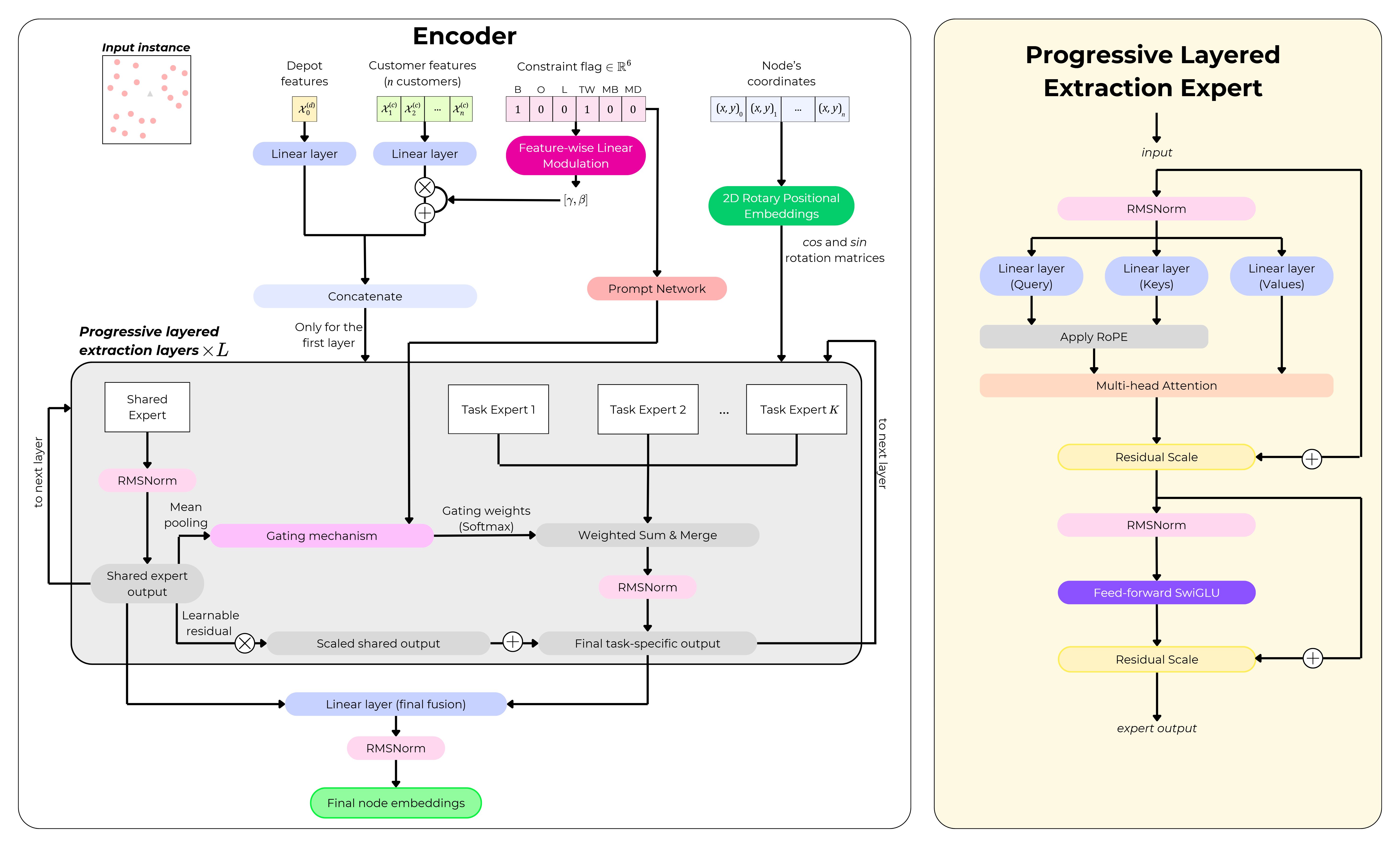}
		\caption{Encoder architecture overview. Left: $L$ PLE layers with one shared expert $E_{\mathrm{shared}}^{(l)}$ and $\kappa$ task-specific experts per layer, coupled by gated residual connections, before fusion into final node embeddings $H^{(L)}$. Right: micro-architecture of one expert block.}
		\label{fig:encoder}
	\end{figure}
	
	Our encoder architecture is summarized in Figure~\ref{fig:encoder}. Input features are projected into a $d$-dimensional latent space, after which a constraint vector $z$ produces a prompt $P$ and FiLM-conditioned embeddings $H^{(0)}$ as specified in Appendix~\ref{app:arch_encoder}. Node embeddings are then refined by $L$ PLE layers.

	Recent encoders for VRPs, such as RouteFinder and CaDA \citep{Berto2025,Li2025} reuse one fully shared stack of $L$ layers across all variants. This design is parameter-efficient, but requires the same weights to encode instances governed by different routing constraints and can increase representation entanglement. Motivated by this, we resort to PLE \citep{Tang2020}, an architecture originally proposed for general MTL to mitigate gradient interference by decomposing shared and task-specific computation through depth. The shared expert provides capacity for routing patterns common to all variants, while the task experts provide additional pathways for constraint-dependent representations. Importantly, specialization is \emph{soft and instance-dependent}, i.e., we do not hard-assign constraints to specific experts. Instead, at every layer we recompute soft routing weights from the constraint prompt and the current shared stream, so the relative expert contributions can change with depth.
	
	Concretely, each PLE layer maintains two parallel streams, a shared stream $H_{\mathrm{shared}}^{(l)}$ and a task stream $H_{\mathrm{task}}^{(l)}$, both initialized from the FiLM-conditioned embeddings $H^{(0)}$. Each PLE layer holds one shared expert $E_{\mathrm{shared}}^{(l)}$ and $\kappa$ task-specific experts $\{E_k^{(l)}\}_{k=1}^{\kappa}$ of identical capacity. Every expert is an MHA Transformer block whose internals are depicted on the right of Figure~\ref{fig:encoder}. At the first layer, the prompt tokens $P$ are concatenated onto the task stream so that task-specific experts can attend jointly to nodes and constraint context.
	
	A forward pass through layer $l$ proceeds as follows. The shared expert first updates the shared stream. Soft gating later attenuates activations, so we apply an outer RMSNorm immediately after this update:
	\begin{equation}
		H_{\mathrm{shared}}^{(l)}
		=
		\mathrm{RMSNorm}\ \big(
		E_{\mathrm{shared}}^{(l)}(H_{\mathrm{shared}}^{(l-1)})
		\big).
	\end{equation}
	The normalized shared representation is later added back into the task stream. Because expert specialization is soft rather than hard-assigned, each layer uses a gate to allocate the $\kappa$ task-specific experts and to decide how strongly the shared stream should enter the task update. We condition this gate on the mean prompt $\bar{P}$, which encodes the active constraint combination, and on a graph-level summary $\mu^{(l)}=\mathrm{mean}(H_{\mathrm{shared}}^{(l)})$ of what the shared expert has extracted at the current depth. A learnable projection $W_p^{(l)}$ first maps $\bar{P}$ to a depth-specific prompt vector. These two vectors are then concatenated and scored by a two-layer MLP with an output of size $K{+}1$:
	\begin{equation}
		g^{(l)} = \mathrm{softmax}\Big(\mathrm{MLP}\big(W_p^{(l)}\bar{P}\,\|\,\mu^{(l)}\big)\Big) \in \mathbb{R}^{\kappa+1}.
	\end{equation}
	The first $\kappa$ components of $g^{(l)}$ weight the task-specific experts, while $g_{\kappa+1}^{(l)}$ scales a shared residual that is added once on top of their mixture. Since this residual lies outside the task mixture, we renormalize the first $\kappa$ weights among themselves:
	\begin{equation}
	\label{eq_weights}
		\tilde{g}_{1:\kappa}^{(l)} = \frac{g_{1:\kappa}^{(l)}}{\sum_{k=1}^{\kappa} g_k^{(l)}},
	\end{equation}
	so that the task-specific expert weights still sum to one and do not shrink when $g_{\kappa+1}^{(l)}$ grows.
	
	We then mix the task-specific expert outputs on $H_{\mathrm{task}}^{(l-1)}$ and add a contribution from the shared stream, gated by $g_{\kappa+1}^{(l)}$ and further scaled by a learnable coefficient $\sigma(\rho^{(l)})$. An outer RMSNorm is applied afterward to keep activation magnitudes stable under this soft mixing:
	\begin{equation}
	\label{eq_task_update}
		H_{\mathrm{task}}^{(l)}
		=
		\mathrm{RMSNorm}\ \Bigl(
		\sum_{k=1}^{\kappa} \tilde{g}_k^{(l)} E_k^{(l)}\!\big(H_{\mathrm{task}}^{(l-1)}\big)
		+
		\sigma\!\big(\rho^{(l)}\big)\,
		g_{\kappa+1}^{(l)}\,
		H_{\mathrm{shared}}^{(l)}
		\Bigr).
	\end{equation}
	The pair $\bigl(H_{\mathrm{shared}}^{(l)},\,H_{\mathrm{task}}^{(l)}\bigr)$ is passed to the next layer. Since $g^{(l)}$ is recomputed at every depth from the prompt and the current shared summary, successive layers can shift emphasis between shared and task-specific computation as specialization progresses.
	
	After $L$ layers, the encoder forms its final representation by concatenating the shared and task \emph{node} streams along the feature dimension, projecting back to $d$ dimensions with a bias-free linear layer, and applying RMSNorm:
	\begin{equation}
	\label{final_rmsnorm}
		H^{(L)}
		=
		\mathrm{RMSNorm}\bigl(
		W_{\mathrm{fuse}}
		\bigl[\,H_{\mathrm{shared}}^{(L)}\,\|\,H_{\mathrm{task}}^{(L)}\,\bigr]
		\bigr)
		\in \mathbb{R}^{(n+1)\times d}.
	\end{equation}
	The resulting embeddings are passed to the decoder to begin the autoregressive process.
	
	\subsection{Preference optimization with local augmented refinement}
	\label{sec_polar_algo}
	
	To train our model, we employ POLAR, a novel algorithm designed for MTL frameworks, which combines PO with an LS algorithm. PO already supplies dense pairwise supervision, but the informativeness of that signal depends on how decisively one decoded tour outranks another. Once convergence stabilizes, many tours for the same instance yield similar costs, so a growing fraction of preference pairs become near-ties and the logistic preference term yields weak gradients. POLAR strengthens the comparison set by applying a lightweight LS before preference labels are formed, so at least one side of each informative pair reflects locally optimized neighborhoods.
	
	Each epoch comprises $I$ instances, sampled in mixed batches of size $B$ \citep{Berto2025}. For each instance $\mathcal{G}_i, \forall i \in \{1, ..., B\}$, the autoregressive decoder first generates $N$ trajectories using POMO's multi-start strategy \citep{Kwon2020}, where each trajectory is initialized from a different customer node. In addition to these $N$ tours, we also generate one extra trajectory greedily, without enforcing a starting node. This gives the model both broad exploratory coverage and a deterministic reference trajectory, which is more likely to provide a better starting point to the LS.
	
	After decoding, the best tour $\tau_i^{\star}$ among the $N{+}1$ candidates is refined by PyVRP's \texttt{LocalSearch} \citep{Wouda2024}, which supports all variants in this work. Importantly, we keep PyVRP's default LS parameters and run it for a single call. After, the refined tour is re-scored under the current policy $p_{\theta}$, with both its reward and log-likelihood replacing those of the unrefined counterpart in the preference set. Preference learning then compares all tours pairwise and increases the probability of preferred trajectories relative to dispreferred ones.
	
	Formally, let $\{\tau_i^{j}\}_{j=0}^{N}$ denote the decoded tours for an instance $\mathcal{G}_i$, with rewards $\{r_i^j\}_{j=0}^{N}$ and log-likelihoods $\{\log p_\theta(\tau_i^{j})\}_{j=0}^{N}$. A Boolean preference label $y_{j,k}^{i}$ indicates whether tour $\tau_i^{j}$ is preferred over tour $\tau_i^{k}$. The PO objective is then calculated as:
	\begin{equation}
		\mathcal{L}_{\text{PO}} = -\frac{1}{B(N+1)^2} \sum_{i=1}^{B} \sum_{j=0, k=0}^{N} y_{j, k}^{i} \log \sigma \bigl(\alpha (\log p_{\theta}(\tau_i^j) - \log p_{\theta}(\tau_i^k))\bigr)
	\end{equation}
	where $\alpha$ is a temperature parameter, and $\sigma$ is a Sigmoid function. Parameters $\theta$ are updated by minimizing $\mathcal{L}_{\mathrm{PO}}$. Algorithm~\ref{alg:polar} summarizes the main training loop. When LS is inactive (during early epochs), line~\ref{alg:polar_ls} is skipped and preference learning proceeds over the raw decoded set.
	
	\begin{algorithm}[t]
		\caption{Preference Optimization with Local-Augmented Refinement (POLAR)}
		\label{alg:polar}
		\begin{algorithmic}[1]
			\Require batch size $B$, epochs $E_{\mathrm{tot}}$, instances per epoch $I$, policy $\theta$, LS start epoch $E_{\mathrm{LS}}$
			\For{epoch = 1, ..., $E_{\mathrm{tot}}$}
			\For{step = 1, ..., $(I // B)$}
			\State Sample mixed batch with $B$ instances of random VRP variants: $\{\mathcal{G}_1, ..., \mathcal{G}_B\}$
			\State Encode each instance to node embeddings
			\State Decode $N+1$ trajectories for each instance: $\tau_i^j, \forall i \in \{1, ..., B\}, \forall j \in \{0, ..., N\}$
			\State Compute rewards $r_i^j = \mathcal{R}(\tau_i^j, \mathcal{G}_i)$ and log-likelihoods $\log p_{\theta}(\tau_i^j), \forall i \in \{1, ..., B\}, \forall j \in \{0, ..., N\}$
			\If{epoch $> E_{\mathrm{LS}}$} \label{alg:polar_ls}
			\State Identify best decoded tour per instance: $j_i^{\star} \leftarrow \arg\max_{j} r_i^{j}$, $\tau_i^{\star} \leftarrow \tau_i^{j_i^{\star}}$
			\State Apply local search: $\tau_i^{\text{LS}} \leftarrow \texttt{LocalSearch}(\tau_i^{\star}), \forall i \in \{1, ..., B\}$
			\If{$\mathcal{R}(\tau_i^{\text{LS}},\mathcal{G}_i) > \mathcal{R}(\tau_i^{\star},\mathcal{G}_i)$}
			\State Replace reward $r_i^{j_i^{\star}}$ and log-likelihood $\log p_{\theta}(\tau_i^{j_i^{\star}})$ with policy scores of $\tau_i^{\text{LS}}$ under $p_\theta$
			\EndIf
			\EndIf
			\State Compute the pairwise preference loss
			$\mathcal{L}_{\text{PO}}$
			\State Update $\theta$ by minimizing $\mathcal{L}_{\text{PO}}$
			\EndFor
			\EndFor
		\end{algorithmic}
	\end{algorithm}
	
	\subsubsection{POLAR design principles}
	
	POLAR is organized around five design principles that jointly target cross-problem generalization, computational efficiency, and scalability, which we summarize below. 
	
	\textbf{Relative, scale-invariant supervision.} POLAR bypasses the need for heterogeneous reward functions or variance-reduction baseline estimations. Instead, the preference signal is obtained directly from pairwise comparisons between decoded trajectories, yielding an $O(N^2)$ supervisory signal per instance. This denser supervision is substantially more informative than the traditional REINFORCE with shared baselines algorithm used in MTL approaches \citep{Kwon2020}, which only evaluates a sampled trajectory against a scalar baseline. Furthermore, because this supervisory signal is relative rather than absolute, the method exhibits inherent scale invariance across diverse VRP variants, mitigating gradient conflicts stemming from tasks with disparate objective function magnitudes.
	
	\textbf{Late activation of refinement.} The LS refinement step is selectively enabled \textit{only in the final 50 epochs out of 300} ($E_{\mathrm{LS}}{=}250$). Empirically, running the LS through the final stages of training performed comparably to running it over the entire training process, while drastically reducing total CPU time. A possible reason is that early training is dominated by mostly coarse policy improvement, when candidate tours are not yet strong enough for refinement to widen meaningful margins.
	
	\textbf{Conservative policy distribution shift.} Each selected tour is refined through a single LS call. This helps to minimize distribution shifts between the policy and refined trajectories. An overly aggressive LS refinement could displace the target tours beyond the current representational capacity of the policy, making the preference targets harder to learn from and the learning process unstable. 
	
	\textbf{CPU-parallel efficiency.} We emphasize algorithmic efficiency by parallelizing the LS execution across the entire batch of instances. This design allows the refinement step to be efficiently offloaded to the CPU, making it scalable across all available processor cores. Consequently, we observed only a small increase in wall-clock time per epoch when the LS was activated compared to when it was disabled, a trade-off heavily outweighed by the substantial gains in solution quality (discussed further in Appendix~\ref{sec:training_efficiency}).
	
	\textbf{Architecture agnosticism.} Finally, POLAR is completely architecture agnostic. The algorithm requires only an encoder with an autoregressive decoder, and a way to re-score tours, meaning it can be integrated into various MTL backbones. We exploit this modularity in the ablation studies presented in Section~\ref{sec:polar_ablation}, demonstrating that POLAR consistently achieves significant performance enhancements across multiple baseline MTL architectures.

	\section{Experiments}
	\label{sec_experiments}
	
	\begin{table}[ht]
		\vskip -0.05in
		\caption{Performance on 1K test instances of 16 seen in-distribution VRPs. Gray cells indicate the best objective and gap among all neural MTL models.}
		\label{tab:main_results}
		\vskip 0.1in
		\begin{center}
			\renewcommand{\arraystretch}{0.82}
			\fontsize{6}{6.5}\selectfont
			\resizebox{\linewidth}{!}{
				\begin{tabular}{ll|cccccc|ll|cccccc}
					\toprule
					\multicolumn{2}{c|}{\multirow{2}{*}{Method}} & \multicolumn{3}{c}{$n=50$} & \multicolumn{3}{c|}{$n=100$} & \multicolumn{2}{c|}{\multirow{2}{*}{Method}} &
					\multicolumn{3}{c}{$n=50$} & \multicolumn{3}{c}{$n=100$} \\
					\cmidrule(lr){3-5} \cmidrule(lr){6-8} \cmidrule(lr){11-13} \cmidrule(lr){14-16}
					& & Obj. & Gap & Time & Obj. & Gap & Time & & & Obj. & Gap & Time & Obj. & Gap & Time \\
					\midrule
					\multirow{9}*{\rotatebox{90}{CVRP}}
					& PyVRP & 10.372 & * & 10.4m & 15.628 & * & 20.8m & \multirow{9}*{\rotatebox{90}{VRPTW}} & PyVRP & 16.031 & * & 10.4m & 25.423 & * & 20.8m \\
					& OR-Tools & 10.572 & 1.907\% & 10.4m & 16.280 & 4.178\% & 20.8m & & OR-Tools & 16.089 & 0.347\% & 10.4m & 25.814 & 1.506\% & 20.8m \\
					& MTPOMO & 10.520 & 1.423\% & 2s & 15.941 & 2.030\% & 8s & & MTPOMO & 16.419 & 2.423\% & 2s & 26.433 & 3.962\% & 9s \\
					& MVMoE & 10.499 & 1.229\% & 3s & 15.888 & 1.693\% & 11s & & MVMoE & 16.400 & 2.298\% & 3s & 26.390 & 3.789\% & 11s \\
					& RouteFinder & 10.502 & 1.257\% & 2s & 15.860 & 1.524\% & 8s & & RouteFinder & 16.341 & 1.933\% & 2s & 26.228 & 3.154\% & 8s \\
					& CaDA & 10.494 & 1.182\% & 2s & 15.870 & 1.578\% & 8s & & CaDA & 16.278 & 1.536\% & 2s & 26.070 & 2.530\% & 8s \\
					& FiLMMeD(CaDA) & 10.468 & 0.924\% & 4s & 15.793 & 1.089\% & 19s & & FiLMMeD(CaDA) & 16.266 & 1.461\% & 4s & 26.029 & 2.365\% & 20s \\
					& MoSES(CaDA) & 10.462 & 0.873\% & 7s & 15.833 & 1.354\% & 24s & & MoSES(CaDA) & 16.262 & 1.435\% & 7s & 26.032 & 2.383\% & 25s \\
					& Ours & \gray{10.446} & \gray{0.717\%} & 3s & \gray{15.751} & \gray{0.817\%} & 14s & & Ours & \gray{16.220} & \gray{1.172\%} & 4s & \gray{25.969} & \gray{2.119\%} & 15s \\
					\midrule
					\multirow{9}*{\rotatebox{90}{OVRP}}
					& PyVRP & 6.507 & * & 10.4m & 9.725 & * & 20.8m & \multirow{9}*{\rotatebox{90}{VRPL}} & PyVRP & 10.587 & * & 10.4m & 15.766 & * & 20.8m \\
					& OR-Tools & 6.553 & 0.686\% & 10.4m & 9.995 & 2.732\% & 20.8m & & OR-Tools & 10.570 & 2.343\% & 10.4m & 16.466 & 5.302\% & 20.8m \\
					& MTPOMO & 6.717 & 3.194\% & 2s & 10.216 & 5.028\% & 8s & & MTPOMO & 10.775 & 1.733\% & 2s & 16.157 & 2.483\% & 8s \\
					& MVMoE & 6.705 & 3.003\% & 3s & 10.177 & 4.617\% & 11s & & MVMoE & 10.753 & 1.525\% & 3s & 16.099 & 2.113\% & 11s \\
					& RouteFinder & 6.682 & 2.658\% & 2s & 10.115 & 3.996\% & 8s & & RouteFinder & 10.747 & 1.485\% & 2s & 16.057 & 1.858\% & 8s \\
					& CaDA & 6.670 & 2.468\% & 2s & 10.121 & 4.045\% & 8s & & CaDA & 10.731 & 1.333\% & 2s & 16.057 & 1.847\% & 8s \\
					& FiLMMeD(CaDA) & 6.649 & 2.148\% & 4s & 10.078 & 3.605\% & 19s & & FiLMMeD(CaDA) & 10.706 & 1.101\% & 4s & 15.988 & 1.419\% & 19s \\
					& MoSES(CaDA) & 6.629 & 1.857\% & 7s & 10.084 & 3.679\% & 24s & & MoSES(CaDA) & 10.704 & 1.083\% & 7s & 16.024 & 1.659\% & 24s \\
					& Ours & \gray{6.612} & \gray{1.596\%} & 4s & \gray{9.998} & \gray{2.785\%} & 15s & & Ours & \gray{10.679} & \gray{0.857\%} & 4s & \gray{15.933} & \gray{1.074\%} & 15s \\
					\midrule
					\multirow{9}*{\rotatebox{90}{VRPB}}
					& PyVRP & 9.687 & * & 10.4m & 14.377 & * & 20.8m & \multirow{9}*{\rotatebox{90}{OVRPTW}} & PyVRP & 10.510 & * & 10.4m & 16.926 & * & 20.8m \\
					& OR-Tools & 9.802 & 1.159\% & 10.4m & 14.933 & 3.853\% & 20.8m & & OR-Tools & 10.519 & 0.078\% & 10.4m & 17.027 & 0.583\% & 20.8m \\
					& MTPOMO & 10.036 & 3.596\% & 2s & 15.102 & 5.052\% & 8s & & MTPOMO & 10.676 & 1.558\% & 2s & 17.442 & 3.022\% & 9s \\
					& MVMoE & 10.007 & 3.292\% & 3s & 15.023 & 4.505\% & 10s & & MVMoE & 10.674 & 1.541\% & 3s & 17.416 & 2.870\% & 12s \\
					& RouteFinder & 9.979 & 3.000\% & 2s & 14.935 & 3.906\% & 8s & & RouteFinder & 10.645 & 1.264\% & 2s & 17.328 & 2.352\% & 9s \\
					& CaDA & 9.960 & 2.800\% & 2s & 14.960 & 4.038\% & 8s & & CaDA & 10.613 & 0.957\% & 2s & 17.226 & 1.751\% & 9s \\
					& FiLMMeD(CaDA) & 9.925 & 2.442\% & 4s & 14.859 & 3.369\% & 19s & & FiLMMeD(CaDA) & 10.611 & 0.940\% & 4s & 17.214 & 1.678\% & 19s \\
					& MoSES(CaDA) & 9.904 & 2.225\% & 7s & 14.901 & 3.668\% & 23s & & MoSES(CaDA) & 10.611 & 0.946\% & 8s & 17.217 & 1.702\% & 26s \\
					& Ours & \gray{9.866} & \gray{1.824\%} & 3s & \gray{14.793} & \gray{2.895\%} & 16s & & Ours & \gray{10.583} & \gray{0.670\%} & 4s & \gray{17.172} & \gray{1.428\%} & 17s \\
					\midrule
					\multirow{9}*{\rotatebox{90}{VRPBL}}
					& PyVRP & 10.186 & * & 10.4m & 14.779 & * & 20.8m & \multirow{9}*{\rotatebox{90}{VRPBLTW}} & PyVRP & 18.361 & * & 10.4m & 29.026 & * & 20.8m \\
					& OR-Tools & 10.331 & 1.390\% & 10.4m & 15.426 & 4.338\% & 20.8m & & OR-Tools & 18.422 & 0.332\% & 10.4m & 29.830 & 2.770\% & 20.8m \\
					& MTPOMO & 10.679 & 4.760\% & 2s & 15.718 & 6.294\% & 8s & & MTPOMO & 19.001 & 2.199\% & 3s & 30.948 & 3.794\% & 9s \\
					& MVMoE & 10.639 & 4.384\% & 3s & 15.642 & 5.771\% & 11s & & MVMoE & 18.983 & 2.097\% & 3s & 30.892 & 3.609\% & 12s \\
					& RouteFinder & 10.569 & 3.713\% & 2s & 15.523 & 5.008\% & 8s & & RouteFinder & 18.910 & 1.713\% & 2s & 30.705 & 2.978\% & 9s \\
					& CaDA & 10.543 & 3.461\% & 2s & 15.525 & 5.001\% & 8s & & CaDA & 18.848 & 1.376\% & 2s & 30.520 & 2.359\% & 9s \\
					& FiLMMeD(CaDA) & 10.512 & 3.152\% & 4s & 15.424 & 4.318\% & 19s & & FiLMMeD(CaDA) & 18.848 & 1.369\% & 4s & 30.499 & 2.287\% & 19s \\
					& MoSES(CaDA) & 10.517 & 3.193\% & 7s & 15.478 & 4.705\% & 24s & & MoSES(CaDA) & 18.858 & 1.425\% & 8s & 30.510 & 2.329\% & 26s \\
					& Ours & \gray{10.454} & \gray{2.579\%} & 4s & \gray{15.345} & \gray{3.781\%} & 16s & & Ours & \gray{18.809} & \gray{1.163\%} & 4s & \gray{30.440} & \gray{2.080\%} & 16s \\
					\midrule
					\multirow{9}*{\rotatebox{90}{VRPBTW}}
					& PyVRP & 18.292 & * & 10.4m & 29.467 & * & 20.8m & \multirow{9}*{\rotatebox{90}{VRPLTW}} & PyVRP & 16.356 & * & 10.4m & 25.757 & * & 20.8m \\
					& OR-Tools & 18.366 & 0.383\% & 10.4m & 29.945 & 1.597\% & 20.8m & & OR-Tools & 16.441 & 0.499\% & 10.4m & 26.259 & 1.899\% & 20.8m \\
					& MTPOMO & 18.649 & 1.938\% & 2s & 30.478 & 3.426\% & 9s & & MTPOMO & 16.832 & 2.877\% & 2s & 26.913 & 4.455\% & 9s \\
					& MVMoE & 18.632 & 1.841\% & 3s & 30.437 & 3.284\% & 12s & & MVMoE & 16.817 & 2.783\% & 3s & 26.866 & 4.272\% & 12s \\
					& RouteFinder & 18.573 & 1.517\% & 2s & 30.249 & 2.641\% & 9s & & RouteFinder & 16.728 & 2.248\% & 2s & 26.706 & 3.645\% & 9s \\
					& CaDA & 18.500 & 1.117\% & 2s & 30.059 & 1.999\% & 9s & & CaDA & 16.669 & 1.879\% & 2s & 26.540 & 2.995\% & 9s \\
					& FiLMMeD(CaDA) & 18.497 & 1.105\% & 4s & 30.042 & 1.938\% & 21s & & FiLMMeD(CaDA) & 16.661 & 1.836\% & 4s & 26.498 & 2.836\% & 20s \\
					& MoSES(CaDA) & 18.495 & 1.095\% & 8s & 30.050 & 1.969\% & 25s & & MoSES(CaDA) & 16.667 & 1.864\% & 8s & 26.493 & 2.824\% & 25s \\
					& Ours & \gray{18.455} & \gray{0.869\%} & 4s & \gray{29.987} & \gray{1.742\%} & 16s & & Ours & \gray{16.612} & \gray{1.534\%} & 4s & \gray{26.416} & \gray{2.513\%} & 16s \\
					\midrule
					\multirow{9}*{\rotatebox{90}{OVRPB}}
					& PyVRP & 6.898 & * & 10.4m & 10.335 & * & 20.8m & \multirow{9}*{\rotatebox{90}{OVRPBL}} & PyVRP & 6.899 & * & 10.4m & 10.335 & * & 20.8m \\
					& OR-Tools & 6.928 & 0.412\% & 10.4m & 10.577 & 2.315\% & 20.8m & & OR-Tools & 6.927 & 0.386\% & 10.4m & 10.582 & 2.363\% & 20.8m \\
					& MTPOMO & 7.105 & 2.973\% & 2s & 10.882 & 5.264\% & 8s & & MTPOMO & 7.112 & 3.053\% & 2s & 10.888 & 5.318\% & 8s \\
					& MVMoE & 7.089 & 2.744\% & 3s & 10.841 & 4.869\% & 11s & & MVMoE & 7.094 & 2.799\% & 3s & 10.847 & 4.929\% & 11s \\
					& RouteFinder & 7.065 & 2.385\% & 2s & 10.774 & 4.233\% & 8s & & RouteFinder & 7.068 & 2.417\% & 2s & 10.778 & 4.266\% & 8s \\
					& CaDA & 7.049 & 2.159\% & 2s & 10.762 & 4.099\% & 8s & & CaDA & 7.051 & 2.166\% & 2s & 10.762 & 4.102\% & 8s \\
					& FiLMMeD(CaDA) & 7.027 & 1.844\% & 4s & 10.695 & 3.456\% & 20s & & FiLMMeD(CaDA) & 7.028 & 1.846\% & 4s & 10.695 & 3.459\% & 20s \\
					& MoSES(CaDA) & 7.034 & 1.942\% & 7s & 10.726 & 3.765\% & 24s & & MoSES(CaDA) & 7.036 & 1.964\% & 7s & 10.724 & 3.743\% & 24s \\
					& Ours & \gray{6.994} & \gray{1.369\%} & 3s & \gray{10.633} & \gray{2.860\%} & 16s & & Ours & \gray{6.996} & \gray{1.386\%} & 3s & \gray{10.633} & \gray{2.858\%} & 16s \\
					\midrule
					\multirow{9}*{\rotatebox{90}{OVRPBLTW}}
					& PyVRP & 11.668 & * & 10.4m & 19.156 & * & 20.8m & \multirow{9}*{\rotatebox{90}{OVRPBTW}} & PyVRP & 11.669 & * & 10.4m & 19.156 & * & 20.8m \\
					& OR-Tools & 11.681 & 0.106\% & 10.4m & 19.305 & 0.767\% & 20.8m & & OR-Tools & 11.682 & 0.109\% & 10.4m & 19.303 & 0.757\% & 20.8m \\
					& MTPOMO & 11.823 & 1.315\% & 3s & 19.658 & 2.602\% & 9s & & MTPOMO & 11.823 & 1.307\% & 3s & 19.656 & 2.592\% & 9s \\
					& MVMoE & 11.816 & 1.249\% & 4s & 19.640 & 2.514\% & 12s & & MVMoE & 11.816 & 1.245\% & 4s & 19.637 & 2.499\% & 13s \\
					& RouteFinder & 11.789 & 1.017\% & 2s & 19.554 & 2.061\% & 9s & & RouteFinder & 11.790 & 1.027\% & 2s & 19.555 & 2.062\% & 9s \\
					& CaDA & 11.760 & 0.771\% & 2s & 19.435 & 1.439\% & 9s & & CaDA & 11.761 & 0.779\% & 2s & 19.436 & 1.441\% & 9s \\
					& FiLMMeD(CaDA) & 11.754 & 0.721\% & 4s & 19.437 & 1.441\% & 22s & & FiLMMeD(CaDA) & 11.754 & 0.719\% & 4s & 19.435 & 1.434\% & 22s \\
					& MoSES(CaDA) & 11.761 & 0.781\% & 8s & 19.440 & 1.470\% & 26s & & MoSES(CaDA) & 11.760 & 0.773\% & 8s & 19.441 & 1.475\% & 26s \\
					& Ours & \gray{11.734} & \gray{0.544\%} & 4s & \gray{19.397} & \gray{1.233\%} & 17s & & Ours & \gray{11.734} & \gray{0.543\%} & 4s & \gray{19.397} & \gray{1.233\%} & 17s \\
					\midrule
					\multirow{9}*{\rotatebox{90}{OVRPL}}
					& PyVRP & 6.507 & * & 10.4m & 9.724 & * & 20.8m & \multirow{9}*{\rotatebox{90}{OVRPLTW}} & PyVRP & 10.510 & * & 10.4m & 16.926 & * & 20.8m \\
					& OR-Tools & 6.552 & 0.668\% & 10.4m & 10.001 & 2.791\% & 20.8m & & OR-Tools & 10.497 & 0.114\% & 10.4m & 17.023 & 0.728\% & 20.8m \\
					& MTPOMO & 6.720 & 3.248\% & 2s & 10.224 & 5.112\% & 8s & & MTPOMO & 10.677 & 1.572\% & 2s & 17.442 & 3.020\% & 9s \\
					& MVMoE & 6.706 & 3.028\% & 3s & 10.184 & 4.693\% & 11s & & MVMoE & 10.677 & 1.564\% & 3s & 17.418 & 2.880\% & 12s \\
					& RouteFinder & 6.683 & 2.680\% & 2s & 10.121 & 4.054\% & 8s & & RouteFinder & 10.646 & 1.267\% & 2s & 17.328 & 2.352\% & 9s \\
					& CaDA & 6.671 & 2.475\% & 2s & 10.122 & 4.052\% & 8s & & CaDA & 10.613 & 0.961\% & 2s & 17.226 & 1.752\% & 9s \\
					& FiLMMeD(CaDA) & 6.650 & 2.157\% & 4s & 10.078 & 3.604\% & 20s & & FiLMMeD(CaDA) & 10.609 & 0.926\% & 4s & 17.214 & 1.678\% & 21s \\
					& MoSES(CaDA) & 6.629 & 1.846\% & 7s & 10.081 & 3.652\% & 24s & & MoSES(CaDA) & 10.611 & 0.940\% & 8s & 17.219 & 1.714\% & 26s \\
					& Ours & \gray{6.612} & \gray{1.582\%} & 4s & \gray{9.997} & \gray{2.782\%} & 15s & & Ours & \gray{10.582} & \gray{0.665\%} & 4s & \gray{17.172} & \gray{1.431\%} & 17s \\
					\bottomrule
			\end{tabular}}
		\end{center}
		\vskip -0.1in
	\end{table}
	
	In this Section, we validate the effectiveness of our method across 16 VRP variants. We also conduct extensive ablation studies and sensitivity analyses to quantify the contribution of each design choice. We further report public-benchmark results in Appendix~\ref{app:public}, zero-shot generalization on 32 unseen VRP variants in Appendix~\ref{app:zeroshot}, and a representation-geometry analysis of encoder embeddings in Appendix~\ref{app:representation_geometry}.
	
	\subsection{Experimental settings}
	
	Unless stated otherwise, we follow the unified protocol established by RouteFinder \citep{Berto2025} and adopted by subsequent cross-problem solvers \citep{Li2025,Correa2026filmmed,Pan2025decomposable}. This alignment keeps instance generation, variant composition, evaluation budgets, and reporting conventions consistent with the current MTL literature, so that observed gains can be attributed to the proposed training algorithm and encoder rather than to mismatched experimental choices.

	We train a separate model for each size $n\in\{50,100\}$ on the 16 single-depot variants formed by combining open routes (O), time windows (TW), distance limits (L), and backhauls (B). Mixed backhauls (MB) and multi-depot (MD) attributes are withheld from this phase and appear only later in the zero-shot evaluation. Training undergoes 300 epochs, with each epoch corresponding to $100{,}000$ instances generated on the fly, sampled in batches of size 256 for $n=50$ and 128 for $n=100$. For the instance generation, we followed the same procedure as \citet{Berto2025}.
	All neural solvers are evaluated with a greedy decoding under the POMO multi-start strategy and $\times 8$ instance augmentation \citep{Kwon2020}. By default, we do not apply POLAR's LS at test time, so that our method incurs no extra inference cost relative to other architectures. We evaluate on the synthetic test sets of \citet{Berto2025}, which comprise 1{,}000 independently sampled instances for each of the 16 VRP variants at $n=50$ and at $n=100$. Instances follow the same attribute and feature distributions used during training, and each model is tested only at the size on which it was trained.

	\textit{Traditional solvers}: We use PyVRP \citep{Wouda2024}, a high-performance implementation of the Hybrid Genetic Search algorithm from \citet{Vidal2022}. We also use the popular Google OR-Tools \citep{ortools}. Both methods solve each instance on a single CPU core, with time limits of 10 seconds for 50-node instances, and 20 seconds for 100-node instances. Since each solver processes thousands of instances, each one is parallelized across 16 CPU cores \citep{Kool2019}.
	\textit{Neural solvers}: We consider here MTPOMO \citep{Liu2024}, MVMoE \citep{Zhou2024}, RouteFinder with its modern Transformer encoder \citep{Berto2025}, CaDA \citep{Li2025}, FiLMMeD(CaDA) \citep{Correa2026filmmed}, and MoSES(CaDA) \citep{Pan2025decomposable}. All methods share the same decoding and augmentation protocol.
	For every method we report the mean objective value, the percentage gap relative to PyVRP, and the total wall-clock time required to solve each 1{,}000-instance test set.

	Our encoder uses an embedding dimension $d=128$, $L=6$ PLE layers with $\kappa{=}3$ task-specific experts, $\eta{=}8$ attention heads, and a feed-forward hidden dimension of 512. Decoder logits are clipped to $[-10,10]$, following \citet{Bello2017}. Training uses the AdamW optimizer \citep{Loshchilov2018} with learning rate $3\times 10^{-4}$, weight decay $10^{-6}$, and automatic mixed precision. The learning rate is multiplied by $0.1$ at epochs 270 and 295, and the PO temperature is $\alpha=0.05$.
	All experiments were conducted on a single machine running Ubuntu 24.04 LTS, equipped with 40 CPU cores, 96 GB of system RAM, and one NVIDIA L40S GPU (48GB VRAM).
	Our code is available at: \url{https://github.com/AJ-Correa/Routing-POLAR}.
	
	\subsection{Main results on synthetic instances}
	
	Table~\ref{tab:main_results} reports the main experimental results for each of the 16 in-distribution variants. Overall, across all state-of-the-art neural solvers, our model attains the best neural objective and gap on all $16$ variants at both instance sizes. Averaging over all 16 variants, the mean gap of our method is $1.192\%$ at $n=50$ and $2.102\%$ at $n=100$. Relative to other published neural baselines, this corresponds to average-gap reductions of $21.3\%$ and $20.1\%$ versus MoSES(CaDA) ($1.515\%$ / $2.631\%$), and of $22.7\%$ and $15.9\%$ versus FiLMMeD(CaDA) ($1.543\%$ / $2.498\%$). Importantly, the performance gains are not achieved by trading generalization on some variants for others, since we observed no performance drop on any of the 16 tasks relative to the competing neural methods. This uniform improvement demonstrates that the aggregate gain is not obtained at the expense of individual training variants.
	
	\subsection{Ablation studies and sensitivity analyses}
	\label{sec:ablations}

	The previous results establish the effectiveness of our full model on the 16 training variants, but they do not quantify how much POLAR and PLE each contribute to overall performance. We therefore conduct a series of controlled ablation studies and sensitivity analyses that isolate one design choice at a time. Unless stated otherwise, the experiments performed in this Section use the 16 standard single-depot variants at $n{=}50$ with the same decoding protocol as the main experiments (POMO multi-start with $\times 8$ augmentation).
	
	\subsubsection{POLAR}
	\label{sec:polar_ablation}
	
	To isolate the contribution of POLAR from architectural improvements, we train four different MTL backbones (MTPOMO, MVMoE, RouteFinder, and our model with PLE) under three training paradigms: REINFORCE with shared baselines (RL), scale-invariant preference optimization (PO) following the implementation of \citet{Correa2026filmmed}, and POLAR.
	
\begin{figure}[t]
		\centering
		\includegraphics[width=0.92\linewidth]{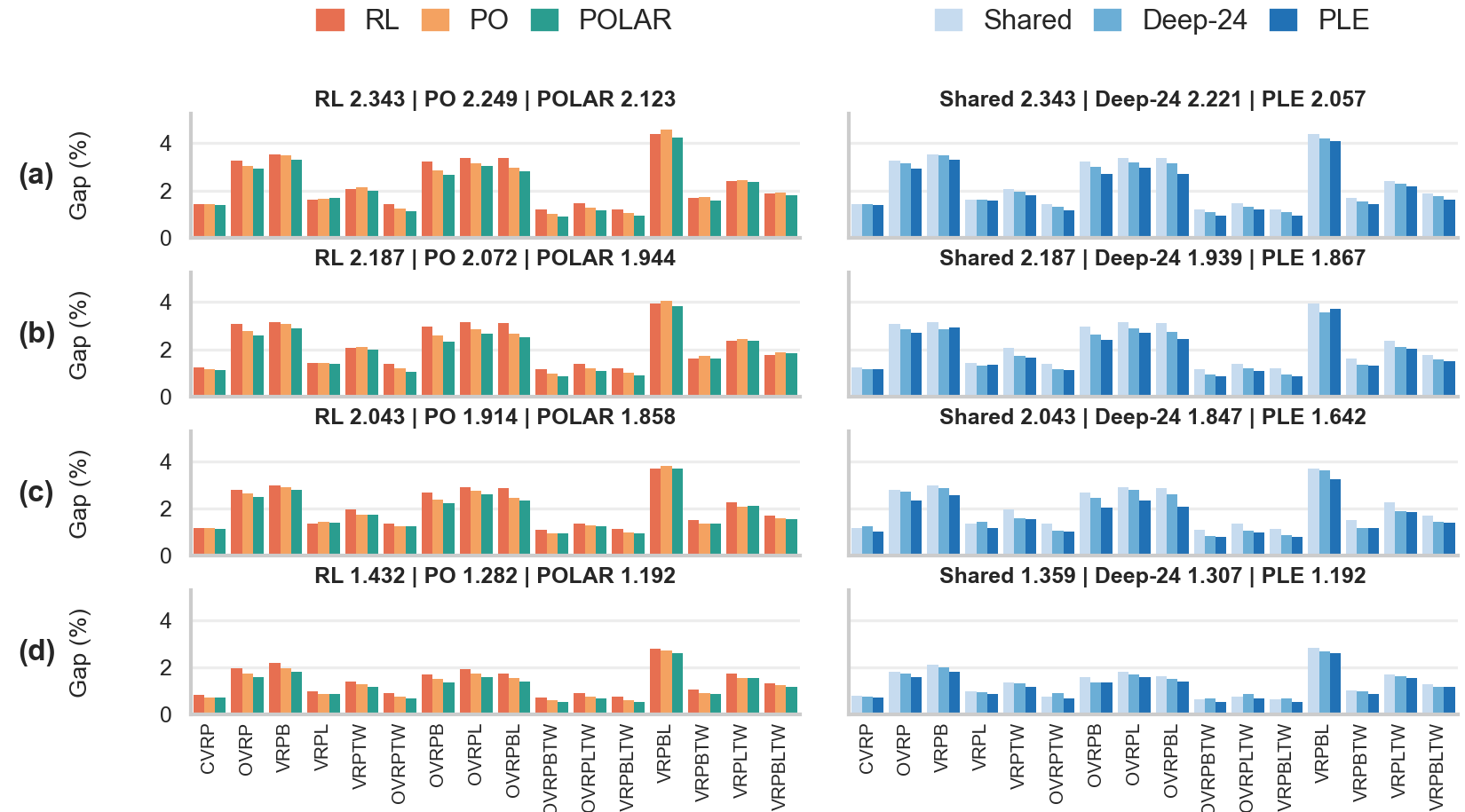}
		\caption{Ablation study on the 16 standard VRP variants. Left: training paradigms (RL, PO, POLAR). Right: encoder settings (Shared, Deep-24, PLE). Rows: (a)~MTPOMO, (b)~MVMoE, (c)~RouteFinder, (d)~our backbone. Panel subtitles report the mean gap (\%) over all variants.}
		\label{fig:polar_paradigm_ablation}
		\label{fig:ple_ablation}
	\end{figure}

	Figure~\ref{fig:polar_paradigm_ablation} (left) reports the gap to PyVRP for every variant--paradigm combination. Across all four backbone architectures, POLAR consistently attains the lowest average gap, improving over both RL and PO, indicating that the training procedure is architecture-agnostic rather than tied to a particular encoder design. On our backbone, the mean gaps are $1.432\%$ (RL), $1.282\%$ (PO), and $1.192\%$ (POLAR). In terms of individual variants, POLAR achieves the best gap on 15/16 variants for MTPOMO, 13/16 for MVMoE, 11/16 for RouteFinder and 16/16 for our model. The remaining cases on the other backbones are near-ties in which PO or RL marginally outperforms POLAR by less than $0.1\%$. This suggests that POLAR does not improve the aggregate at the cost of individual variants. Even when LS yields little or no gain over PO or RL on a given task, performance remains comparable, with no pronounced degradation on any single variant.
	
	\FloatBarrier
	
	\noindent
	\begin{minipage}[t]{0.55\linewidth}
	\vspace{0pt}
	\raggedright
	\textbf{Temperature sensitivity.}
	\label{sec:polar_alpha_sensitivity}
	The paradigm comparison establishes that POLAR improves over both RL and PO. We next examine whether this advantage depends on the temperature $\alpha$ in $\mathcal{L}_{\mathrm{PO}}$, which governs how strongly log-probability differences are converted into preference gradients. To isolate the training procedure from encoder capacity, we sweep $\alpha \in \{0.01, 0.03, 0.05, 0.07, 0.09\}$ for PO and POLAR on our backbone with PLE disabled. Table~\ref{tab:alpha_sensitivity} reports the resulting mean gaps.
	\end{minipage}\hfill
	\begin{minipage}[t]{0.42\linewidth}
		\vspace{0pt}
		\centering
		\footnotesize
		\setlength{\abovecaptionskip}{0pt}
		\setlength{\belowcaptionskip}{8pt}
		\captionof{table}{Sensitivity of mean performance gap (\%) to preference temperature $\alpha$. Best per column in gray.}
		\label{tab:alpha_sensitivity}
		\setlength{\tabcolsep}{6pt}
		\begin{tabular}{ccc}
			\toprule
			$\alpha$ & PO & POLAR \\
			\midrule
			0.01 & 1.513 & 1.449 \\
			0.03 & 1.502 & 1.433 \\
			0.05 & \gray{1.441} & \gray{1.359} \\
			0.07 & 1.507 & 1.378 \\
			0.09 & 1.688 & 1.531 \\
			\bottomrule
		\end{tabular}
	\end{minipage}
	
	\vspace{0.5em}
	\noindent Both paradigms are comparatively stable in the mid-range of the sweep and degrade at the extremes, most sharply at $\alpha{=}0.09$. POLAR is uniformly better than PO at every temperature, and the performance difference widens as $\alpha$ increases, peaking at $\alpha{=}0.09$. This pattern suggests that the LS-refined anchor keeps preference learning more robust when pairwise rankings become overly sharp. The best setting is shared by both methods at $\alpha{=}0.05$, which we adopt in all subsequent experiments.
	
	\noindent
	\begin{minipage}[t]{0.52\linewidth}
	\vspace{0pt}
	\raggedright
	\textbf{Refinement subset size.}
	\label{sec:polar_refinement_subset}
	We argue in Section~\ref{sec_polar_algo} that POLAR refines only the best decoded tour before preference labels are formed, in order to keep CPU overhead modest during training. To verify whether this choice sacrifices solution quality, we vary the number of locally refined tours per instance from $k{=}1$ to $k{=}5$. Table~\ref{tab:topk_refinement} reports the resulting mean performance gap and average wall-clock time per training epoch.
	\end{minipage}\hfill
	\begin{minipage}[t]{0.45\linewidth}
		\vspace{0pt}
		\centering
		\footnotesize
		\setlength{\abovecaptionskip}{0pt}
		\setlength{\belowcaptionskip}{8pt}
		\captionof{table}{Effect of refining the top-$k$ decoded tours in POLAR. Best per column in gray.}
		\label{tab:topk_refinement}
		\setlength{\tabcolsep}{4pt}
		\begin{tabular}{ccc}
			\toprule
			$k$ & Mean gap (\%) & Avg.\ epoch time \\
			\midrule
			1 & \gray{1.359} & \gray{6m 44s} \\
			2 & 1.367 & 7m 04s \\
			3 & 1.395 & 7m 19s \\
			4 & 1.417 & 7m 36s \\
			5 & 1.461 & 7m 50s \\
			\bottomrule
		\end{tabular}
	\end{minipage}
	
	\vspace{0.5em}
	\noindent The mean gap increases monotonically with $k$, so enlarging the refined subset does not improve solution quality, while training time grows roughly linearly. This is expected given how refinement is used in training. Each polished tour replaces its corresponding POMO candidate in place, and preference learning still compares all tours in that same sample pairwise. Once the best decoded tour has been refined, it likely drives most of the informative comparisons, since lower-ranked candidates are unlikely to overtake it after being refined by the LS. Thus, refining more tours adds little new supervision beyond what the top tour already provides.
	Additionally, refining many tours can be mildly harmful, as it increases the number of near-tie preferred pairs among comparable refined solutions and weakens the sharp contrasts that PO relies on for learning. Together with the distribution-shift concern noted in Section~\ref{sec_polar_algo}, where overly aggressive refinement can push targets beyond the policy's current support, these effects explain why larger $k$ raises the gap despite the extra CPU work. Refining a single tour therefore remains the most favorable trade-off between training cost and solution quality, hence our choice of setting $k{=}1$.
	
	\subsubsection{Progressive layered extraction}
	\label{sec:ple_ablation}
	To assess the individual effects of the PLE encoder, we train MTPOMO, MVMoE, RouteFinder, and our model under three encoder settings: (i) \textit{Shared}, a fully shared encoder with $L{=}6$ layers and no PLE; (ii) \textit{PLE}, the same backbone with Progressive Layered Extraction; and (iii) \textit{Deep-24}, a fully shared encoder with depth increased to $L{=}24$. Because each PLE layer runs one shared expert together with $\kappa$ task-specific experts, PLE also increases capacity relative to Shared at the same depth. Deep-24 therefore matches the number of expert blocks in our default PLE configuration ($L{=}6$ with one shared expert and $\kappa{=}3$ task-specific experts), separating raw capacity from the shared/task decomposition. The right column of Figure~\ref{fig:ple_ablation} reports the per-variant performance gap for each of the three settings.
	
	Across all four backbone architectures, PLE attains the lowest average performance gap. Deep-24 improves over Shared on every backbone, which confirms that extra depth helps to some extent. Even so, Deep-24 remains strictly worse than PLE in every case, so extra capacity alone does not explain the observed gains. The consistent advantage of PLE over both baselines indicates that the improvement comes from progressive shared/task decomposition rather than from simply training a larger encoder. Additionally, since these gains appear across four different model architectures, it also indicates that PLE acts as an architecture-agnostic mechanism rather than being tied to a single model design.
	
	\noindent
	\begin{minipage}[t]{0.50\linewidth}
	\vspace{0pt}
	\raggedright
	\textbf{Number of task experts.}
	We next analyze the sensitivity on the number of task-specific experts $\kappa$ used in each PLE layer. Table~\ref{tab:ple_experts} reports the mean performance gap together with the average wall-clock time per training epoch before and after local search is activated. Overall, increasing $\kappa$ steadily improves average performance, but the returns diminish while training time grows considerably in both regimes. Thus, setting $\kappa{=}3$ offers a favorable balance, reaching near-best solution quality at a lower average epoch time than $\kappa{=}4$.
	\end{minipage}\hfill
	\begin{minipage}[t]{0.47\linewidth}
		\vspace{0pt}
		\centering
		\footnotesize
		\setlength{\abovecaptionskip}{0pt}
		\setlength{\belowcaptionskip}{8pt}
		\captionof{table}{Sensitivity of mean gap (\%) and epoch time to the number of PLE task experts $\kappa$. Best gap in gray.}
		\label{tab:ple_experts}
		\setlength{\tabcolsep}{3.5pt}
		\begin{tabular}{cccc}
			\toprule
			$\kappa$ & Mean gap (\%) & Epoch w/o LS & Epoch w/ LS \\
			\midrule
			1 & 1.286 & 3m 29s & 6m 41s \\
			2 & 1.221 & 3m 36s & 6m 50s \\
			3 & 1.192 & 3m 43s & 7m 01s \\
			4 & \gray{1.176} & 3m 51s & 7m 11s \\
			\bottomrule
		\end{tabular}
	\end{minipage}
	
	\section{Conclusion}
	\label{sec_conclusion}
	
	In this paper, we addressed two persistent limitations of existing multi-task VRP solvers. First, we introduced POLAR, which strengthens preference supervision by refining the best decoded tour before pairwise comparison, preserving more informative preference margins as the policy converges. Second, we proposed a PLE encoder that separates common routing structure from constraint-dependent representations through progressively coupled shared and task-specific experts, mitigating representation entanglement in fully shared models. Through extensive experiments, we demonstrated that the combination of POLAR and PLE elevates the current state-of-the-art across numerous VRP variants. Ablation studies confirm the individual contribution of both components. POLAR improves four distinct MTL backbones over both RL and PO training, while PLE outperforms both typical fully shared encoders and a parameter-matched deeper shared model. One limitation that remains open is handling harder and more complex real-world constraints. In the future, we aim to significantly expand the current benchmark to hundreds of different VRP variants, providing a much broader MTL test suite under richer constraint combinations.

\subsubsection*{Acknowledgments}
This research is sponsored by national funds through FCT -- Funda\c{c}\~ao para a Ci\^encia e a Tecnologia (FCT), Portugal, through doctoral grant 2025.00622.BD and projects UID/00285/2025, and LA/P/0112/2020. This work was further supported by the European Union -- NextGenerationEU, through the Portuguese Republic's Recovery and Resilience Plan Partnership Agreement [project C645808870-00000067], within the scope of the project PRODUTECH R3 -- ``Agenda Mobilizadora da Fileira das Tecnologias de Produ\c{c}\~ao para a Reindustrializa\c{c}\~ao'', Total project investment: 166.988.013,71 Euros; Total Grant: 97.111.730,27 Euros; and by the European Regional Development Fund (ERDF) through the Operational Program for Competitiveness and Internationalization (COMPETE 2020) under the project POCI-01-0247-FEDER-046102 (PRODUTECH4S\&C). Finally, this work was also supported by FCT's High Performance Computing contest under the grant reference 2026.07669.CPCA.

\bibliography{cas-refs}
\bibliographystyle{iclr2026_conference}

\clearpage
\appendix
	\section{Related work}
	\label{sec_related_work}
	\label{app:related}
	
	\subsection{Neural-based CO}
	
	Existing works on neural-based VRP solvers can be generally classified into three categories. \textit{Construction-based} methods learn to build solutions from scratch in an end-to-end manner. Early work used Pointer Networks \citep{Vinyals2015}, a sequence-to-sequence architecture that uses a 'pointer' mechanism to dynamically select input elements as outputs, allowing the output size to adapt to the input. While showing promising results on the Traveling Salesman Problem (TSP), Pointer Networks rely on Supervised Learning and therefore require (near-)optimal labels that are expensive to obtain at scale for NP-hard routing problems. Subsequent works replaced Supervised Learning with RL, removing the dependency on exact labels \citep{Bello2017,Nazari2018}. A major architectural milestone followed with the Attention-based model of \citet{Kool2019}, which has since become the standard for routing works. Building on this design, \citet{Kwon2020} introduced Policy Optimization with Multiple Optima (POMO), advancing neural-based solvers by exploiting the symmetries inherent in CO problems. POMO introduced a REINFORCE with shared baselines algorithm that is specifically tailored to CO problems. By generating multiple trajectories for the same instance, with each one starting from a different initial node, it computes a low-variance baseline that stabilizes training and improves robustness to local minima. Furthermore, the authors introduced an $\times8$ instance augmentation technique that applies rotations and reflections to the 2D Euclidean map of node coordinates. Subsequent approaches were mostly developed as improvements over the previously cited works \citep{Xin2021,Kim2022,Chalumeau2023,Drakulic2023,Luo2023,Correa2026}. 
	
	\textit{Improvement-based} methods take a different route. Instead of autoregressively constructing a tour, these methods learn policies to select and control search operators that iteratively refine an incumbent solution \citep{Chen2019neurewriter,Lu2020,Ma2023}. A notable example is NeuroLKH \citep{Xin2021neurolkh}, a hybrid approach combining a Sparse Graph Network with the LKH heuristic \citep{Helsgaun2000}. It employs supervised edge scoring alongside unsupervised node penalties to guide the search process. Ultimately, while improvement-based approaches typically offer higher solution quality than construction-based methods, they tend to be more computationally expensive, which limits their suitability for real-time applications. Lastly, a recent third line of work has focused on \textit{Divide-and-Conquer} methods, which learn to decompose instances into smaller sub-problems \citep{Cheng2023select,Hou2023generalize,Ye2024glop,Zheng2024udc}. While Divide-and-Conquer methods scale well to large instances, they risk compounding errors across sub-problem boundaries and impose additional implementation complexity from coordinating multiple neural and heuristic components. Across all three paradigms, autoregressive construction remains the preferred one in MTL settings because it offers the best balance between solution quality, inference speed, and compatibility with variant-conditioned representations.
	
	\subsection{Multi-task learning for VRPs}
	
	Recently, the landscape of neural-based CO has shifted significantly towards MTL to handle the huge variety of existing VRP variants. \citet{Liu2024} established the foundation of MTL research with MTPOMO, the first cross-problem model for VRPs, which uses an attribute composition block to solve 16 distinct variants. This was later extended by \citet{Zhou2024} through MVMoE, a mixture-of-experts approach with hierarchical gating to improve model capacity and generalization. Building on this, \citet{Berto2025} introduced RouteFinder, a foundation model framework for VRPs that integrates modern Transformer architectural innovations. RouteFinder also differs from previous works by introducing mixed-batch training, allowing instances from multiple variants to be sampled within the same batch. After, \citet{Li2025} proposed CaDA, a dual-attention model that incorporates a top-$k$ sparse attention branch into the encoder. By making each node attend to its $k$ most relevant neighbors, CaDA effectively encodes localized spatial relationships within the graph, resulting in better generalization than previous architectures.
	
	Various subsequent works have focused on refining representations, decoder capacity, and scalability. For instance, ReLD \citep{Huang2025} and PoMtVRS \citep{Meng2026} augment light decoders with state attribute mapping and feed-forward residual blocks, significantly improving generalization. \citet{Zheng2025} introduced MTL-KD, a knowledge distillation method which distills a student model from single-task teachers to improve cross-size generalization. More recently, \citet{Gui2026} proposed Chain-of-Context Learning, which refines static node embeddings during autoregressive decoding. While these contributions improve decoding capacity, they leave the encoder largely shared across variants and thus, do not directly resolve interference among heterogeneous constraints.
	
	Therefore, a distinct and, for cross-problem settings, more central line of research targets representation disentanglement across variants. \citet{Wang2026} proposed SPSM, a soft parameter sharing model that utilizes two separate Multi-Head Attention (MHA) modules to disentangle the commonalities and peculiarities across VRP variants. \citet{Pan2025decomposable} introduced MoSES, a mixture-of-specialized-experts solver that decomposes VRP variants into conditionally independent basis states through a novel State-Decomposable Markov Decision Process formulation. MoSES utilizes specialized LoRA experts to learn individual constraints, which are then dynamically integrated via an adaptive gating mechanism. More recently, \citet{Correa2026filmmed} introduced FiLMMeD, which applies a lightweight feature-wise linear modulation mechanism to condition internal representations based on the active constraint vector at each instance.
	
	Despite recent advances, previous methods remain limited in some regards. First, although MoSES \citep{Pan2025decomposable} reports significant improvements over prior state-of-the-art models, it requires training multiple experts for each constraint type, making its training cost prohibitive. Second, while SPSM improves generalization over different baselines, its reliance on a single task-specific module to capture all variant-specific features leaves room for finer disentanglement. To address this, we propose a PLE encoder which progressively merges representations from shared and task-specific experts. Our approach uses a single MTL model, avoiding the need to train separate experts. Furthermore, the architecture is entirely agnostic to the underlying model, yielding consistent performance gains across various baselines with no performance drops on any problem variant.
	
	\subsection{Alternative training paradigms}
	
	Beyond architectural limitations, existing MTL frameworks are also constrained by their training paradigms. Most rely on RL, which suffers from the fundamental limitations discussed in Section~\ref{sec:intro}. To reduce the reliance on unstable scalar rewards, recent neural CO research has actively explored training paradigms outside of RL. Self-labeling methods, such as the ones introduced by \citet{Corsini2024} and \citet{luo2025boosting}, reduce dependence on scalar signals by sampling multiple solutions for each individual instance and treating the best sampled tour as a pseudo-label. While outperforming other classic RL algorithms, they tend to be sample inefficient since only one decoded solution is used, with the rest being discarded. Additionally, they may suffer from confirmation bias, as the model can reinforce its own suboptimal patterns. 
	
	In an alternative line of work, other studies have applied Direct Preference Optimization \citep{Rafailov2023} to CO problems. Unlike traditional RL, PO reframes learning as discriminating better from worse solutions rather than maximizing a reward. For instance, \citet{Liao2025} introduced BOPO, a novel algorithm that constructs $k$ preference pairs anchored to the best sampled solution. While BOPO achieves better generalization than other algorithms, its objective-guided scaling factor still depends on raw objective values, potentially reintroducing sensitivity to heterogeneous objective function scales across different problem variants. Alternatively, \citet{Pan2025} presented a more generic PO formulation that constructs all $(\substack{N \\ 2})$ possible preference pairs between sampled solutions using only relative rankings, rendering the approach inherently scale-invariant. The authors further proposed refining decoded solutions with an LS during a separate fine-tuning phase. However, the proposed design exhaustively applied the LS to a vast number of solutions, significantly increasing the computational burden per epoch. In addition, their approach is demonstrated only on single-task CVRP and TSP settings, leaving open the question on how to achieve both computational efficiency and broad multi-task applicability. While FiLMMeD \citep{Correa2026filmmed} recently set the standard for using PO instead of RL for MTL models, it does not fully exploit the potential of high-quality pairwise comparisons, since it lacks mechanisms to actively enhance the preference pairs themselves.
	
	Our work bridges these gaps by introducing POLAR, a novel training algorithm for MTL models that combines scale-invariant preference supervision with a cross-problem LS. Unlike \citet{Pan2025}, which uses an LS to refine a subset of solutions for every instance, POLAR selectively refines only the best solution, avoiding excessive computational burden. Crucially, we demonstrate that refining solely the top solution achieves similar performance to refining a larger top-$k$ subset. Beyond efficiency, POLAR is also highly versatile, supporting \textit{all VRP variants solved in current MTL models}. Additionally, POLAR is \textit{architecture-agnostic}, making it compatible with virtually any underlying MTL model. Finally, we address a critical limitation overlooked by prior PO methods: \textit{as the policy converges, generated solutions become increasingly similar, yielding preference pairs with diminishing quality gaps that can stall learning}. By systematically enhancing anchor solution quality through a targeted search procedure, POLAR sustains informative comparisons, enabling a better generalization.

\section{Learning to solve VRPs}
\label{app:learning}
	
	In this study, as in previous works \citep{Kool2019,Liu2024}, we parametrize the policy $\pi_{\theta}$ using an attention-based neural network. Solution construction proceeds autoregressively via an encoder-decoder architecture: an encoder first maps the problem instance $\mathcal{G}$ into $d$-dimensional node embeddings, which an MHA decoder then uses to build the solution sequentially. This process can be formalized as a Markov Decision Process (MDP) defined by the tuple $(\mathcal{S}, \mathcal{A}, \mathcal{T}, \mathcal{R})$:
	
	\textbf{State space $\mathcal{S}$}: At timestep $t$, the agent observes a state $s_t \in \mathcal{S}$ comprising the encoded node embeddings and a context vector $\mathcal{D}_t$ that summarizes the partial solution. 
	
	\textbf{Action space $\mathcal{A}$}: Given state $s_t$, the agent selects an action $a_t \in \mathcal{A}_t$, where $\mathcal{A}_t$ denotes the set of feasible next nodes at timestep $t$. Actions may correspond to visiting an unserved customer or returning to the departing depot to close the current route. Here, the policy $\pi_\theta$ generates a probability distribution over $\mathcal{A}_t$. To ensure feasibility, a mask is applied, masking all infeasible nodes by setting their logits to $-\infty$. During training, actions are sampled from this distribution, while during inference, they are selected greedily (i.e., the node with the highest probability is always chosen).
	
	\textbf{Transition function $\mathcal{T}$}: After selecting $a_t$, the transition function $\mathcal{T}$ updates the state to $s_{t+1}$, reflecting the changes caused by that action. In particular, it updates key attributes of the partial solution, such as the vehicle's location, remaining capacity, and the current route’s length and duration. This cycle of state observation, action selection and transition then continues until a full feasible solution $\tau$ is constructed.
	
	\textbf{Reward function $\mathcal{R}$}: Once a complete solution $\tau$ is constructed, the agent receives a reward $\mathcal{R}(\tau, \mathcal{G}) = -\text{cost}(\tau)$ representing the negative total traveled distance across all routes in $\tau$. In the end, the probability of generating this solution under policy $\pi_\theta$ can be factorized as $p_{\theta}(\tau | \mathcal{G}) = \Pi_{t=1}^{T} \ p_{\theta} (a_t | a_{t-1}, ..., a_1, \mathcal{G})$, where $T$ is the total number of steps taken during the decoding process.

	\section{Attention}
\label{app:attention}
\label{sec:attention}
	
	Attention is the mechanism through which MTL models exchange information among embedded nodes, allowing the model to capture long-range spatial and contextual dependencies across an instance. In this work, we employ the classic self-attention mechanism \citep{Vaswani2017} within both the encoder and decoder architectures. Let $X \in \mathbb{R}^{B \times N \times d}$ denote a batch of node embeddings, where $B$ is the batch size, $N$ is the number of nodes in the VRP instance, and $d$ is the embedding dimension. Learned linear projections first map the input embeddings into queries, keys, and values:
	\begin{equation}
		Q = XW_Q, \qquad K = XW_K, \qquad V = XW_V,
	\end{equation}
	where $W_Q, W_K,$ and $W_V$ are trainable parameter matrices.
	
	The resulting tensors are reshaped into $\eta$ attention heads of dimension $d_h$, yielding head-specific representations:
	\begin{equation}
		Q,K,V \in \mathbb{R}^{B \times \eta \times N \times d_h}.
	\end{equation}
	Each head independently computes attention scores and aggregates information across nodes. The outputs of all heads are then concatenated and projected back to the model dimension $d$. Then, for each head, the model calculates an affinity matrix, defined as:
	\begin{equation}
		S_i = \frac{Q_iK_i^\top}{\sqrt{d_h}},
	\end{equation}
	where $Q_i, K_i,$ and $V_i$ denote the query, key, and value representations associated with head $i$. An optional mask can be applied, adding large negative values to exclude invalid actions before normalization (in the context of VRPs, infeasible nodes). Attention weights are then obtained through a row-wise softmax:
	\begin{equation}
		\mathrm{MHA}(Q_i,K_i,V_i) = \mathrm{softmax}(S_i)V_i.
	\end{equation}
	
	This mechanism allows every query node to attend to all keys, enabling global information exchange and dense contextualization across the input sequence.

	\section{Detailed architecture overview}
\label{app:architecture}

This appendix specifies the encoder components that are not unique to PLE, together with the full autoregressive decoder. The PLE routing itself is specified in Section~\ref{sec:model_architecture}.

	\subsection{Encoder}
	\label{app:arch_encoder}

	The model initiates by processing an input instance $\mathcal{G}$, which consists of depot $\mathcal{X}_{0}$ and customer features $\{\mathcal{X}_{i}\}_{i=1}^{n}$. The depot features include its 2D coordinates, and distance limits, while customer nodes are characterized by their 2D coordinates, linehaul and backhaul demands, time windows, and service times. Both sets of features are projected into a $d$-dimensional latent space through separate linear layers.
	
	To make representations variant-aware, active constraints are encoded using a Boolean vector $z\in\{0,1\}^6$ indicating the presence of open routes, time windows, distance limits, backhauls, mixed backhauls, and multi-depots (the last two of which are unseen attributes, used only in zero-shot settings). These flags enter the encoder in two complementary ways. First, a prompt network \citep{Li2025} maps $z$ to a prompt vector $P\in\mathbb{R}^{6\times d}$. In parallel, we condition customer embeddings through a feature-wise linear modulation module \citep{Correa2026filmmed}. A multi-layer perceptron, composed of three linear layers with two intermediate ReLU activations, produces per-dimension scale and shift parameters:
	\begin{equation}
		[\gamma\,|\,\beta] = f_{\mathrm{FiLM}}(z) \in \mathbb{R}^{2d}.
	\end{equation}
	The customer embeddings are then modulated as:
	\begin{equation}
		\mathrm{FiLM}(h_i \mid \gamma, \beta) = \gamma \odot h_i + \beta, \ \forall i \in \{1, ..., n\}
	\end{equation}
	Identical raw attributes can therefore be interpreted differently under distinct constraint combinations before attention is applied.
	
	Having formed the FiLM-conditioned embeddings $H^{(0)}$ and the constraint prompt $P$, the encoder applies the $L$ PLE layers introduced in Section~\ref{sec:model_architecture}. Each shared or task-specific expert in that stack is an MHA Transformer block with the following micro-architecture (right side of Figure~\ref{fig:encoder}): RMSNorm \citep{Zhang2019rms}, multi-head self-attention, a residual update scaled by $(2L)^{-1/2}$ following \citet{Radford2019}, a second RMSNorm, a SwiGLU feed-forward network \citep{Shazeer2020}, and another residual with the same depth-aware scale. Within every expert block, we also apply a 2D Rotary Positional Embedding \citep{Su2024} to queries and keys using node coordinates, so attention affinities depend on relative spatial displacement rather than absolute position.

	\subsection{Decoder}
	\label{app:arch_decoder}
	
	After the encoding stage, the decoder initializes its attention keys and values using the final node embeddings $H^{(L)}$. The decoder operates autoregressively, i.e., at each step, it conditions on the current partial solution to select the next feasible node. Let $h_{\tau_{t-1}}^{(L)}$ be the embedding of the current node. We augment it with a compact dynamic state vector $\mathcal{D}_t$, which contains critical information about the current partial solution. This includes the remaining linehaul and backhaul vehicle's capacities, the elapsed time of the current route, the length of the current route, and a Boolean indicating whether the current problem has open routes or not. The decoder query is built from concatenating $\mathcal{D}_t$ with $h_{\tau_{t-1}}^{(L)}$ and projected into multi-head query vectors. In parallel, the encoded node embeddings are linearly projected once to form the cached key and value tensors used throughout decoding, as well as the single-head representation employed in the final scoring step. This keeps decoding lightweight, since the keys and values do not need to be recomputed at every timestep. The query then attends to these cached keys and values through multi-head attention, with infeasible nodes masked out, yielding one output vector per head.
	
	Before these heads are projected back to dimension $d$, we apply a preference-gated mechanism \citep{Meng2026}. A sigmoid gate predicted from $[h_{\tau_{t-1}}^{(L)}\,\|\,\mathcal{D}_t]$ produces one scalar per head, and each head's output is multiplied by its corresponding scalar. The gated heads are then linearly combined into a glimpse vector. This glimpse is refined with a residual block that gives the lightweight decoder substantially more capacity. Specifically, we first map $\mathcal{D}_t$ into the embedding space through a learned linear transformation and add it to $h_{\tau_{t-1}}^{(L)}$. The resulting vector is then added to the glimpse and normalized with RMSNorm. A SwiGLU feed-forward network follows, again under an RMSNorm residual connection, producing the final decoder representation $\hat{h}_t$. This sequence of operations significantly improves the expressiveness of an otherwise shallow decoder, with a negligible increase in computational overhead compared to heavy decoder architectures.
	
	Finally, the decoder computes compatibility scores between $\hat{h}_t$ and all encoded nodes through a single-head dot-product operation. Let $\mathcal{A}_t$ denote the set of feasible nodes at step $t$, and let $h_i^{(L)}$ be the final encoded embedding of node $v_i$. The compatibility score of node $v_i$ is then defined as:
	\begin{equation}
		u_{t,i} =
		\begin{cases}
			\xi \tanh\left( \dfrac{(\hat{h}_t)^{\top} h_i^{(L)}}{\sqrt{d}} \right), & \text{if } v_i \in \mathcal{A}_t, \\
			-\infty, & \text{otherwise},
		\end{cases}
	\end{equation}
	where $\xi$ is a logit clipping constant. Here, the compatibilities of infeasible nodes are set to $-\infty$. Lastly, the probability of selecting node $v_i$ at step $t$ is then obtained through:
	\begin{equation}
		p_{t,i} = \mathrm{softmax}(u_{t,i}) = \frac{\exp(u_{t,i})}{\sum_j \exp(u_{t,j})}.
	\end{equation}
	This distribution is used to sample actions during training, whereas during inference, nodes are selected greedily.
	
	\section{Further discussion}
\label{app:further}
	
	The ablation studies in Section~\ref{sec:ablations} confirm the effectiveness of each proposed contribution, but do not examine whether the learned behavior is consistent with the design motivations established earlier in the paper. We therefore report complementary analyses of POLAR's preference supervision and the computation performed inside the PLE encoder. As with the ablation studies, all experiments use $n{=}50$ and the same $\times 8$ instance augmentation with multi-start decoding strategy.
	
	\subsection{POLAR from scratch}
	Section~\ref{sec_polar_algo} motivates activating the LS only in the final 50 epochs of training in order to reduce total wall-clock time, since early decoded tours may not yet be from good enough regions of the solution search space for refinement to yield informative preference margins. To test this empirically, we compare our default schedule ($E_{\mathrm{LS}}{=}250$) against training with LS enabled from epoch 1 on the same backbone. Under late activation, average performance gap is $1.192\%$ and the full 300-epoch run requires about 21h~15m of wall-clock time. In contrast, running POLAR from scratch reaches a mean gap of $1.125\%$ but needs about 35h~9m in total, since every epoch undergoes local refinement, inducing much more CPU overhead. The gain is therefore relatively small in solution quality ($0.067$ percentage points) relative to a $65\%$ increase in training time. Late activation thus preserves most of the POLAR benefit while keeping the additional CPU overhead manageable.
	
	\begin{table}[t]
		\centering
		\caption{Preference-margin statistics during training for PO and POLAR. Gray cells mark the more informative regime (higher $\Delta$ or $\Delta_{\theta}$ and lower near-tie fraction).}
		\label{tab:preference_margin}
		\setlength{\tabcolsep}{8pt}
		\footnotesize
		\begin{tabular}{lcc}
			\toprule
			Metric & PO & POLAR \\
			\midrule
			Mean preferred-pair gap $\Delta$ & 0.519 & \gray{0.562} \\
			Cost near-tie fraction @ $\varepsilon{=}0.1\%$ & 2.36\% & \gray{2.13\%} \\
			Cost near-tie fraction @ $\varepsilon{=}0.5\%$ & 11.28\% & \gray{10.29\%} \\
			Cost near-tie fraction @ $\varepsilon{=}1\%$ & 21.66\% & \gray{19.91\%} \\
			Cost near-tie fraction @ $\varepsilon{=}5\%$ & 74.79\% & \gray{71.83\%} \\
			\hline
			\addlinespace
			Mean policy gap $\Delta_{\theta}$ & 0.295 & \gray{0.366} \\
			Policy near-tie fraction @ $\varepsilon{=}0.1\%$ & 0.32\% & \gray{0.25\%} \\
			Policy near-tie fraction @ $\varepsilon{=}0.5\%$ & 1.59\% & \gray{1.25\%} \\
			Policy near-tie fraction @ $\varepsilon{=}1\%$ & 3.18\% & \gray{2.49\%} \\
			Policy near-tie fraction @ $\varepsilon{=}5\%$ & 15.69\% & \gray{12.37\%} \\
			\bottomrule
		\end{tabular}
	\end{table}
	
	\subsection{Preference-margin analysis}
	\label{sec:polar_margin_analysis}
	
	Throughout the paper, we argue that PO becomes less informative as policies improve because many decoded tours for the same instance obtain similar costs, so an increasing share of preference pairs are near-ties and the logistic term in $\mathcal{L}_{\mathrm{PO}}$ yields weak gradients. POLAR targets this limitation by refining the best decoded tour with an LS before preference labels are assigned, so that preferred comparisons reflect locally optimized neighborhoods.
	
	To quantify how well this mechanism operates in practice, we monitor preference supervision during training under standard PO and under POLAR over epochs $251$--$300$ (i.e., after the LS is activated in POLAR). At each training batch, we form the usual pairwise preference matrix over the decoded solutions. Let $y_{j,k}{=}1$ when tour $\tau^j$ is preferred over $\tau^k$. We first record the mean preferred-pair cost gap:
	\begin{equation}
		\Delta = \mathbb{E}\!\left[\, c(\tau^j) - c(\tau^k) \;\middle|\; y_{j,k} = 1 \,\right],
	\end{equation}
	together with the fraction of preferred pairs that are cost near-ties.
	For a preferred pair $(j,k)$, we define the relative cost gap as:
	\begin{equation}
		\rho_{j,k} = \frac{\lvert c(\tau^j) - c(\tau^k)\rvert}{c(\tau^j)} \times 100 \%,
	\end{equation}
	where $c(\tau^j)$ is the cost of the preferred tour.
	A pair is a cost near-tie at tolerance $\varepsilon$ if $\rho_{j,k} < \varepsilon$, with $\varepsilon \in \{0.1, 0.5, 1, 5\}\%$.
	Larger $\Delta$ and lower near-tie rates indicate more decisive cost contrasts in the preference set.
	
	Cost gaps alone can overstate the benefit of POLAR, since refining the best tour with LS tends to widen $|c(\tau^j)-c(\tau^k)|$ even when the policy does not rank the preferred tour more confidently. Hence, we also evaluate the same preferred pairs in the policy space after the refined tour is re-scored under $p_{\theta}$. With PO temperature $\alpha$, the mean absolute policy gap is:
	\begin{equation}
		\Delta_{\theta} = \mathbb{E}\!\left[\,\lvert \alpha\bigl(\log p_{\theta}(\tau^j)-\log p_{\theta}(\tau^k)\bigr) \rvert \;\middle|\; y_{j,k} = 1\,\right].
	\end{equation}
	Policy near-ties are defined analogously from the relative gap in log-likelihoods at the same $\varepsilon$ values.
	
	Table~\ref{tab:preference_margin} shows that POLAR widens $\Delta$ and reduces near-ties relative to PO, as expected once LS is applied. The policy-space metrics provide further evidence. Relative to PO, POLAR increases $\Delta_{\theta}$ across all reported $\varepsilon$. Preferred tours are therefore assigned more decisive log-likelihoods under the current policy after re-scoring, which supports the claim that POLAR improves preference supervision beyond the trivial effect of lowering costs with the LS.
	
	\subsection{Contributions of the PLE experts}
	\label{sec:ple_contributions}
	The PLE encoder is intended to preserve routing information that is useful across variants while providing task experts with the capacity to model constraint-dependent behavior. To examine whether this division is reflected in the trained model, we measure how much each expert contributes to the embedding updates at every encoder layer.
	
	More specifically, we pass mixed batches of randomly sampled VRP variants through the trained encoder and, at layer~$l$, form the four gated terms that enter the update in Equation~\eqref{eq_task_update}:
	the three task-expert outputs, each scaled by $\tilde{g}_k^{(l)}$ (Equation~\ref{eq_weights}), and the shared residual, scaled by $\sigma(\alpha^{(l)})\,g_{K+1}^{(l)}$.
	For each term, we compute the mean node-wise $\ell_2$ norm and express it as a percentage of the total norm across all four terms, averaged over instances. Table~\ref{tab:ple_contributions} reports, at each layer, the percentage contributed by the shared and task experts immediately before the final RMSNorm from Equation~\eqref{final_rmsnorm}.
	
	\begin{table}[t]
		\centering
		\caption{Relative contribution of each PLE expert to the embedding update.}
		\label{tab:ple_contributions}
		\setlength{\tabcolsep}{5.5pt}
		\footnotesize
		\begin{tabular}{ccccc}
			\toprule
			Layer & Task 1 & Task 2 & Task 3 & Shared \\
			\midrule
			1 & 57.5\% & 19.0\% & 12.9\% & 10.7\% \\
			2 & 15.6\% & 39.6\% & 28.3\% & 16.5\% \\
			3 & 18.1\% & 11.3\% & 41.7\% & 28.9\% \\
			4 & 13.6\% & 36.7\% & 30.2\% & 19.5\% \\
			5 & 24.8\% & 28.2\% & 22.1\% & 24.8\% \\
			6 & 26.0\% & 34.1\% & 39.9\% & 0.02\% \\
			\bottomrule
		\end{tabular}
	\end{table}
	
	Overall, the task pathways remain active throughout the encoder, jointly accounting for $71.1$--$89.3\%$ of the contribution magnitude in the first five layers. The shared pathway supplies the remaining $10.7$--$28.9\%$, maintaining common routing information alongside the task computation. We also observe that expert emphasis changes according to depth. Task~1 contributes most in the first layer, Task~2 in layers 2, 4, and 5, and Task~3 in layers 3 and 6. Hence, no single task expert dominates the full stack of layers. In the final layer, the task update combines the three experts at $26.0\%$, $34.1\%$, and $39.9\%$, while the shared residual falls to $0.02\%$. This shift suggests that common routing information has already been established by the preceding updates, allowing the last embedding update to concentrate on constraint-dependent refinement immediately before the shared and task streams are fused for decoding.

	\section{Computational analysis}
\label{app:compute}
	\label{sec:training_efficiency}
	
	In this subsection, we quantify the computational cost required by the proposed training procedure and encoder design through various metrics. We compare four configurations: POLAR+PLE (POLAR training algorithm with the PLE encoder), POLAR (the same training objective with a fully shared encoder), PLE (REINFORCE with the PLE encoder), and Shared (REINFORCE with a fully shared encoder, the usual design choice across MTL literature). Table~\ref{tab:computational_analysis} compares the cost of each component and of their combination.
	
	\begin{table}[t]
		\centering
		\caption{Training cost on one NVIDIA L40S GPU. Average epoch time and total time follow the 300-epoch training schedule. Memory is the peak allocated GPU memory during training.}
		\label{tab:computational_analysis}
		\setlength{\tabcolsep}{10pt}
		\footnotesize
		\begin{tabular}{lccccc}
			\toprule
			Method & $n$ & Parameters & Avg.\ epoch time & Total time & Peak GPU memory \\
			\midrule
			POLAR+PLE & 50  & 6.98M & 4m 18s  & 21h 30m & 15.66 GB \\
			POLAR     & 50  & 1.84M & 4m 16s  & 21h 20m & 15.03 GB \\
			PLE       & 50  & 6.98M & 3m 37s  & 18h 05m & 13.73 GB \\
			Shared    & 50  & 1.84M & 3m 36s  & 18h 00m & 12.52 GB \\
			\hline
			\addlinespace
			POLAR+PLE & 100 & 6.98M & 14m 49s & 74h 04m & 27.00 GB \\
			POLAR     & 100 & 1.84M & 13m 42s & 68h 30m & 23.55 GB \\
			PLE       & 100 & 6.98M & 12m 15s & 61h 14m & 24.03 GB \\
			Shared    & 100 & 1.84M & 11m 14s & 56h 11m & 21.33 GB \\
			\bottomrule
		\end{tabular}
	\end{table}
	
	Under $\kappa=3$, PLE adds its extra capacity entirely within the encoder, raising the parameter count from 1.84M to 6.98M. The encoder processes each instance once, whereas the autoregressive decoder repeatedly scores actions throughout tour construction. This difference explains why the $3.79\times$ parameter increase produces a much smaller change in training time. At $n{=}50$, PLE changes the average epoch time (averaged over all 300 epochs) by less than $1\%$. At $n{=}100$, it increases average epoch time by about $8.1\%$ with POLAR and $9.0\%$ with REINFORCE, while peak GPU memory rises by $14.6\%$ and $12.7\%$, respectively.
	
	Therefore, local refinement is the main source of additional computational cost of our work. For the PLE model, activating POLAR raises the average epoch time from 4m 16s to 4m 18s at $n{=}50$ and from 12m 15s to 14m 49s at $n{=}100$. The same pattern holds for the shared encoder, where the corresponding times increase from 3m 36s to 3m 37s and from 11m 14s to 13m 42s. It is important to note that POLAR applies local refinement only during the final 50 of 300 epochs, thus, although an epoch with refinement is slower, it accounts for only one sixth of the total training schedule.

	\begin{table}[H]
		\centering
		\tiny
		\caption{Results on CVRPLib instances from the Set-X~\citep{Uchoa2017}. Gray-shaded cells indicate the best objective and gap among all neural models.}
		\label{tab:setX_results}
		\resizebox{\linewidth}{!}
		{\begin{tabular}{ll|cccccccccccccccc}
				\toprule
				\multicolumn{2}{c|}{Set-X} & \multicolumn{2}{c}{MTPOMO} & \multicolumn{2}{c}{MVMoE} & \multicolumn{2}{c}{RouteFinder} & \multicolumn{2}{c}{CaDA} & \multicolumn{2}{c}{FiLMMeD(CaDA)} & \multicolumn{2}{c}{MoSES(CaDA)} & \multicolumn{2}{c}{Ours} \\
				\midrule
				Instance & Opt. & Cost & Gap & Cost & Gap & Cost & Gap & Cost & Gap & Cost & Gap & Cost & Gap & Cost & Gap \\
				\midrule
				X-n101-k25 & 27591 & 29470 & 6.810\% & 29076 & 5.382\% & 29048 & 5.281\% & 28944 & 4.904\% & 29095 & 5.451\% & 29110 & 5.505\% & \gray{28810} & \gray{4.418\%} \\
				X-n106-k14 & 26362 & 28029 & 6.323\% & 27443 & 4.101\% & 27159 & 3.023\% & 27042 & 2.579\% & 26826 & 1.760\% & 27051 & 2.614\% & \gray{26801} & \gray{1.665\%} \\
				X-n110-k13 & 14971 & \gray{15100} & \gray{0.862\%} & 15327 & 2.378\% & 15314 & 2.291\% & 15229 & 1.723\% & 15203 & 1.549\% & 15332 & 2.411\% & 15163 & 1.282\% \\
				X-n115-k10 & 12747 & 13433 & 5.382\% & 13475 & 5.711\% & 13338 & 4.636\% & \gray{13060} & \gray{2.455\%} & 13244 & 3.899\% & 13085 & 2.652\% & 13071 & 2.542\% \\
				X-n120-k6 & 13332 & 14051 & 5.393\% & 13782 & 3.375\% & 13765 & 3.248\% & 13678 & 2.595\% & \gray{13552} & \gray{1.650\%} & 13619 & 2.153\% & 13563 & 1.733\% \\
				X-n125-k30 & 55539 & 59015 & 6.259\% & 58200 & 4.791\% & 58570 & 5.457\% & 57748 & 3.977\% & 57815 & 4.098\% & \gray{57620} & \gray{3.747\%} & 58051 & 4.523\% \\
				X-n129-k18 & 28940 & 30176 & 4.271\% & \gray{29334} & \gray{1.361\%} & 29457 & 1.786\% & 29500 & 1.935\% & 29397 & 1.579\% & 29620 & 2.350\% & 29349 & 1.413\% \\
				X-n134-k13 & 10916 & 11707 & 7.246\% & 11462 & 5.002\% & 11624 & 6.486\% & 11652 & 6.742\% & 11367 & 4.132\% & 11573 & 6.019\% & \gray{11296} & \gray{3.481\%} \\
				X-n139-k10 & 13590 & 14058 & 3.444\% & 14099 & 3.745\% & 13812 & 1.634\% & 13940 & 2.575\% & 13929 & 2.494\% & 13877 & 2.112\% & \gray{13787} & \gray{1.450\%} \\
				X-n143-k7 & 15700 & 16626 & 5.898\% & 16349 & 4.134\% & 16257 & 3.548\% & 16189 & 3.115\% & 16107 & 2.592\% & \gray{15980} & \gray{1.783\%} & 16160 & 2.930\% \\
				X-n148-k46 & 43448 & 46648 & 7.365\% & 45893 & 5.627\% & \gray{45026} & \gray{3.632\%} & 45606 & 4.967\% & 45704 & 5.193\% & 45600 & 4.953\% & 45220 & 4.078\% \\
				X-n153-k22 & 21220 & 23514 & 10.811\% & 23661 & 11.503\% & 23478 & 10.641\% & 23142 & 9.057\% & 23019 & 8.478\% & 23310 & 9.849\% & \gray{22976} & \gray{8.275\%} \\
				X-n157-k13 & 16876 & 17922 & 6.198\% & 17439 & 3.336\% & 17315 & 2.601\% & 17295 & 2.483\% & 17270 & 2.335\% & 17317 & 2.613\% & \gray{17154} & \gray{1.647\%} \\
				X-n162-k11 & 14138 & 14616 & 3.381\% & 14705 & 4.010\% & 14664 & 3.720\% & 14704 & 4.003\% & 14713 & 4.067\% & 14677 & 3.812\% & \gray{14519} & \gray{2.695\%} \\
				X-n167-k10 & 20557 & 21662 & 5.375\% & 21504 & 4.607\% & 21425 & 4.222\% & \gray{21078} & \gray{2.534\%} & 21180 & 3.029\% & 21384 & 4.023\% & 21232 & 3.284\% \\
				X-n172-k51 & 45607 & 48960 & 7.352\% & 47883 & 4.990\% & 48162 & 5.602\% & 48198 & 5.681\% & 47543 & 4.245\% & 48145 & 5.565\% & \gray{47466} & \gray{4.076\%} \\
				X-n176-k26 & 47812 & 51989 & 8.736\% & 52117 & 9.004\% & 51501 & 7.716\% & \gray{51120} & \gray{6.919\%} & 51711 & 8.155\% & 51612 & 7.948\% & 51314 & 7.325\% \\
				X-n181-k23 & 25569 & 26572 & 3.923\% & 26456 & 3.469\% & 26097 & 2.065\% & 26262 & 2.710\% & 26024 & 1.780\% & 26143 & 2.245\% & \gray{25993} & \gray{1.658\%} \\
				X-n186-k15 & 24145 & 25236 & 4.519\% & 25151 & 4.166\% & 25153 & 4.175\% & 25345 & 4.970\% & 25197 & 4.357\% & 25246 & 4.560\% & \gray{25026} & \gray{3.649\%} \\
				X-n190-k8 & 16980 & 18369 & 8.180\% & 19078 & 12.356\% & 17871 & 5.247\% & 17882 & 5.312\% & 17812 & 4.900\% & 17569 & 3.469\% & \gray{17513} & \gray{3.139\%} \\
				X-n195-k51 & 44225 & 48310 & 9.237\% & 46974 & 6.216\% & 47396 & 7.170\% & 46723 & 5.648\% & \gray{46400} & \gray{4.919\%} & 47479 & 7.358\% & 46823 & 5.875\% \\
				X-n200-k36 & 58578 & 62041 & 5.912\% & 61627 & 5.205\% & 61139 & 4.372\% & 61010 & 4.152\% & \gray{60833} & \gray{3.850\%} & 61089 & 4.287\% & 60883 & 3.935\% \\
				X-n204-k19 & 19565 & 20652 & 5.556\% & 20584 & 5.208\% & 20531 & 4.937\% & 20735 & 5.980\% & 20491 & 4.733\% & 20420 & 4.370\% & \gray{20285} & \gray{3.680\%} \\
				X-n209-k16 & 30656 & 32333 & 5.470\% & 32358 & 5.552\% & 31876 & 3.980\% & 32184 & 4.984\% & 32248 & 5.193\% & 32053 & 4.557\% & \gray{31813} & \gray{3.774\%} \\
				X-n214-k11 & 10856 & 11699 & 7.765\% & 11597 & 6.826\% & 11668 & 7.480\% & 11748 & 8.217\% & 11659 & 7.397\% & 11716 & 7.922\% & \gray{11315} & \gray{4.228\%} \\
				X-n219-k73 & 117595 & 121980 & 3.729\% & 124434 & 5.816\% & 120344 & 2.338\% & 120011 & 2.055\% & 122996 & 4.593\% & 119710 & 1.799\% & \gray{119248} & \gray{1.406\%} \\
				X-n223-k34 & 40437 & 43381 & 7.280\% & 42694 & 5.582\% & 42251 & 4.486\% & 42273 & 4.540\% & 42178 & 4.305\% & 42128 & 4.182\% & \gray{41872} & \gray{3.549\%} \\
				X-n228-k23 & 25742 & 28523 & 10.803\% & 28033 & 8.900\% & 28699 & 11.487\% & 27821 & 8.076\% & 27523 & 6.919\% & 27724 & 7.699\% & \gray{27400} & \gray{6.441\%} \\
				X-n233-k16 & 19230 & 20644 & 7.353\% & 20656 & 7.415\% & 20761 & 7.962\% & 20285 & 5.486\% & 20538 & 6.801\% & 20623 & 7.244\% & \gray{20216} & \gray{5.127\%} \\
				X-n237-k14 & 27042 & 30047 & 11.112\% & 29772 & 10.095\% & 29595 & 9.441\% & 30282 & 11.981\% & 29952 & 10.762\% & 29518 & 9.156\% & \gray{28092} & \gray{3.883\%} \\
				X-n242-k48 & 82751 & 88179 & 6.559\% & 87497 & 5.735\% & 85704 & 3.569\% & 85813 & 3.700\% & 85576 & 3.414\% & 85643 & 3.495\% & \gray{85310} & \gray{3.092\%} \\
				X-n247-k50 & 37274 & 41610 & 11.633\% & 40973 & 9.924\% & 40642 & 9.036\% & \gray{39918} & \gray{7.093\%} & 40444 & 8.504\% & 40736 & 9.288\% & 40427 & 8.459\% \\
				\midrule
				\multicolumn{2}{c|}{Avg. Gap ($n<251$)} & \multicolumn{2}{c}{6.567\%} & \multicolumn{2}{c}{5.798\%} & \multicolumn{2}{c}{5.102\%} & \multicolumn{2}{c}{4.786\%} & \multicolumn{2}{c}{4.598\%} & \multicolumn{2}{c}{4.742\%} & \multicolumn{2}{c}{\gray{3.710\%}} \\
				\midrule
				X-n251-k28 & 38684 & 41211 & 6.532\% & 41330 & 6.840\% & \gray{40127} & \gray{3.730\%} & 40359 & 4.330\% & 40247 & 4.040\% & 40290 & 4.152\% & 40215 & 3.958\% \\
				X-n256-k16 & 18839 & 20400 & 8.286\% & 20559 & 9.130\% & 19994 & 6.131\% & 20372 & 8.137\% & 20086 & 6.619\% & 20068 & 6.524\% & \gray{19902} & \gray{5.643\%} \\
				X-n261-k13 & 26558 & 28741 & 8.220\% & 28524 & 7.403\% & 28510 & 7.350\% & 28833 & 8.566\% & 28631 & 7.805\% & 28577 & 7.602\% & \gray{28061} & \gray{5.659\%} \\
				X-n266-k58 & 75478 & 84617 & 12.108\% & 82048 & 8.705\% & 79832 & 5.769\% & 80115 & 6.144\% & 80247 & 6.318\% & 80036 & 6.039\% & \gray{79250} & \gray{4.997\%} \\
				X-n270-k35 & 35291 & 38146 & 8.090\% & 38333 & 8.620\% & 37382 & 5.925\% & 37674 & 6.752\% & 37360 & 5.862\% & \gray{36923} & \gray{4.624\%} & 37063 & 5.021\% \\
				X-n275-k28 & 21245 & 24688 & 16.206\% & 25021 & 17.774\% & 24187 & 13.848\% & 24482 & 15.237\% & 24347 & 14.601\% & 24312 & 14.436\% & \gray{22047} & \gray{3.775\%} \\
				X-n280-k17 & 33503 & 36677 & 9.474\% & 36636 & 9.351\% & 36653 & 9.402\% & 36081 & 7.695\% & 35844 & 6.987\% & 35494 & 5.943\% & \gray{35058} & \gray{4.641\%} \\
				X-n284-k15 & 20226 & 22474 & 11.114\% & 22583 & 11.653\% & 22154 & 9.532\% & 22295 & 10.229\% & 21916 & 8.355\% & 22071 & 9.122\% & \gray{21464} & \gray{6.121\%} \\
				X-n289-k60 & 95151 & 104159 & 9.467\% & 102202 & 7.410\% & 100418 & 5.535\% & 99739 & 4.822\% & 100808 & 5.945\% & 100080 & 5.180\% & \gray{99700} & \gray{4.781\%} \\
				X-n294-k50 & 47161 & 52769 & 11.891\% & 50886 & 7.898\% & 50637 & 7.370\% & 49929 & 5.869\% & 49980 & 5.977\% & \gray{49877} & \gray{5.759\%} & 49950 & 5.914\% \\
				X-n298-k31 & 34231 & 37652 & 9.994\% & 37344 & 9.094\% & 37163 & 8.565\% & 36993 & 8.069\% & 36497 & 6.620\% & 37068 & 8.288\% & \gray{36303} & \gray{6.053\%} \\
				X-n303-k21 & 21736 & 23556 & 8.373\% & 23263 & 7.025\% & 23442 & 7.849\% & 23748 & 9.257\% & 23295 & 7.174\% & 23548 & 8.336\% & \gray{22814} & \gray{4.960\%} \\
				X-n308-k13 & 25859 & 28736 & 11.126\% & 28518 & 10.283\% & 28326 & 9.540\% & 28913 & 11.810\% & 28219 & 9.125\% & 28440 & 9.981\% & \gray{27654} & \gray{6.941\%} \\
				X-n313-k71 & 94043 & 102253 & 8.730\% & 100620 & 6.994\% & 99564 & 5.871\% & 98899 & 5.164\% & 99511 & 5.814\% & 98931 & 5.198\% & \gray{98597} & \gray{4.842\%} \\
				X-n317-k53 & 78355 & 82587 & 5.401\% & 83632 & 6.735\% & 80690 & 2.980\% & 80542 & 2.791\% & 81804 & 4.402\% & 80472 & 2.702\% & \gray{80366} & \gray{2.567\%} \\
				X-n322-k28 & 29834 & 32593 & 9.248\% & 33497 & 12.278\% & 32658 & 9.466\% & 33206 & 11.303\% & 32959 & 10.474\% & 32541 & 9.074\% & \gray{31822} & \gray{6.664\%} \\
				X-n327-k20 & 27532 & 30646 & 11.310\% & 30603 & 11.154\% & 29784 & 8.180\% & 30953 & 12.426\% & 30280 & 9.980\% & 30089 & 9.287\% & \gray{29374} & \gray{6.690\%} \\
				X-n331-k15 & 31102 & 34734 & 11.678\% & 33636 & 8.147\% & 34048 & 9.472\% & 34578 & 11.176\% & 34200 & 9.960\% & 34014 & 9.363\% & \gray{33014} & \gray{6.148\%} \\
				X-n336-k84 & 139111 & 152846 & 9.873\% & 149229 & 7.273\% & 146620 & 5.398\% & 146707 & 5.460\% & 149982 & 7.815\% & 146465 & 5.286\% & \gray{145666} & \gray{4.712\%} \\
				X-n344-k43 & 42050 & 46619 & 10.866\% & 46947 & 11.646\% & 44914 & 6.811\% & 45571 & 8.373\% & 45044 & 7.121\% & 44746 & 6.411\% & \gray{44530} & \gray{5.898\%} \\
				X-n351-k40 & 25896 & 29243 & 12.925\% & 28373 & 9.565\% & 28236 & 9.036\% & 28059 & 8.353\% & 28101 & 8.515\% & 28130 & 8.627\% & \gray{27327} & \gray{5.526\%} \\
				X-n359-k29 & 51505 & 55778 & 8.296\% & 56165 & 9.048\% & 55122 & 7.023\% & 55183 & 7.141\% & 54884 & 6.561\% & 55158 & 7.093\% & \gray{53645} & \gray{4.155\%} \\
				X-n367-k17 & 22814 & 26132 & 14.544\% & 25588 & 12.159\% & 25522 & 11.870\% & 25534 & 11.923\% & 25094 & 9.994\% & 25489 & 11.725\% & \gray{24798} & \gray{8.696\%} \\
				X-n376-k94 & 147713 & 156857 & 6.190\% & 156546 & 5.980\% & 151975 & 2.885\% & 151390 & 2.489\% & 152539 & 3.267\% & 151614 & 2.641\% & \gray{150302} & \gray{1.753\%} \\
				X-n384-k52 & 65940 & 73705 & 11.776\% & 73570 & 11.571\% & 70471 & 6.871\% & 70611 & 7.084\% & 70044 & 6.223\% & 70479 & 6.884\% & \gray{69723} & \gray{5.737\%} \\
				X-n393-k38 & 38260 & 43533 & 13.782\% & 44638 & 16.670\% & 41552 & 8.604\% & 42934 & 12.216\% & 41929 & 9.590\% & 42192 & 10.277\% & \gray{40781} & \gray{6.589\%} \\
				X-n401-k29 & 66154 & 71565 & 8.179\% & 71787 & 8.515\% & 69430 & 4.952\% & 69875 & 5.625\% & 69596 & 5.203\% & 69991 & 5.800\% & \gray{68605} & \gray{3.705\%} \\
				X-n411-k19 & 19712 & 23869 & 21.089\% & 23139 & 17.385\% & 22849 & 15.914\% & 23521 & 19.323\% & 23198 & 17.685\% & 22768 & 15.503\% & \gray{21976} & \gray{11.485\%} \\
				X-n420-k130 & 107798 & 122761 & 13.881\% & 116362 & 7.944\% & 117418 & 8.924\% & 115012 & 6.692\% & 115837 & 7.457\% & 116853 & 8.400\% & \gray{114745} & \gray{6.444\%} \\
				X-n429-k61 & 65449 & 74261 & 13.464\% & 74158 & 13.307\% & 70164 & 7.204\% & 70969 & 8.434\% & 70546 & 7.792\% & 70617 & 7.896\% & \gray{69297} & \gray{5.879\%} \\
				X-n439-k37 & 36391 & 41165 & 13.119\% & 42161 & 15.856\% & 39752 & 9.236\% & 41149 & 13.075\% & 39887 & 9.606\% & 39697 & 9.085\% & \gray{38707} & \gray{6.364\%} \\
				X-n449-k29 & 55233 & 60162 & 8.924\% & 60015 & 8.658\% & 60634 & 9.779\% & 61144 & 10.702\% & 60812 & 10.101\% & 60723 & 9.940\% & \gray{58941} & \gray{6.713\%} \\
				X-n459-k26 & 24139 & 29543 & 22.387\% & 29100 & 20.552\% & 27347 & 13.290\% & 28267 & 17.101\% & 27435 & 13.654\% & 27510 & 13.965\% & \gray{26286} & \gray{8.894\%} \\
				X-n469-k138 & 221824 & 252031 & 13.618\% & 245581 & 10.710\% & 238904 & 7.700\% & 237548 & 7.089\% & 243469 & 9.758\% & 237001 & 6.842\% & \gray{235153} & \gray{6.009\%} \\
				X-n480-k70 & 89449 & 101314 & 13.265\% & 100121 & 11.931\% & 95032 & 6.242\% & 95466 & 6.727\% & 95167 & 6.392\% & 95211 & 6.442\% & \gray{93486} & \gray{4.513\%} \\
				X-n491-k59 & 66483 & 77536 & 16.625\% & 75226 & 13.151\% & 72618 & 9.228\% & 71702 & 7.850\% & 72031 & 8.345\% & 71730 & 7.892\% & \gray{70484} & \gray{6.018\%} \\
				\midrule
				\multicolumn{2}{c|}{Avg. Gap ($251 \leq n<501$)} & \multicolumn{2}{c}{11.390\%} & \multicolumn{2}{c}{10.512\%} & \multicolumn{2}{c}{7.986\%} & \multicolumn{2}{c}{8.762\%} & \multicolumn{2}{c}{8.087\%} & \multicolumn{2}{c}{7.842\%} & \multicolumn{2}{c}{\gray{5.680\%}} \\
				\midrule
				X-n502-k39 & 69226 & 75711 & 9.368\% & 77033 & 11.278\% & 71908 & 3.874\% & 72655 & 4.953\% & 72108 & 4.163\% & \gray{71682} & \gray{3.548\%} & 71771 & 3.676\% \\
				X-n513-k21 & 24201 & 34910 & 44.250\% & 32858 & 35.771\% & 28542 & 17.937\% & 29422 & 21.573\% & 28844 & 19.185\% & 29139 & 20.404\% & \gray{26816} & \gray{10.805\%} \\
				X-n524-k153 & 154593 & 176491 & 14.165\% & 171734 & 11.088\% & 174150 & 12.651\% & 168181 & 8.790\% & \gray{165183} & \gray{6.849\%} & 172580 & 11.635\% & 167408 & 8.290\% \\
				X-n536-k96 & 94846 & 109897 & 15.869\% & 106031 & 11.793\% & 103242 & 8.852\% & 102355 & 7.917\% & 102551 & 8.123\% & 101712 & 7.239\% & \gray{101376} & \gray{6.885\%} \\
				X-n548-k50 & 86700 & 110984 & 28.009\% & 104240 & 20.231\% & 100850 & 16.321\% & 102318 & 18.014\% & 101100 & 16.609\% & 101918 & 17.552\% & \gray{91541} & \gray{5.584\%} \\
				X-n561-k42 & 42717 & 55936 & 30.946\% & 53110 & 24.330\% & 49133 & 15.020\% & 50287 & 17.721\% & 48537 & 13.624\% & 49363 & 15.558\% & \gray{46399} & \gray{8.620\%} \\
				X-n573-k30 & 50673 & 60884 & 20.151\% & 62033 & 22.418\% & 56048 & 10.607\% & 55353 & 9.236\% & 54571 & 7.692\% & 55058 & 8.654\% & \gray{53796} & \gray{6.163\%} \\
				X-n586-k159 & 190316 & 226245 & 18.879\% & 212545 & 11.680\% & 205654 & 8.059\% & 204649 & 7.531\% & 208415 & 9.509\% & 204848 & 7.636\% & \gray{201272} & \gray{5.757\%} \\
				X-n599-k92 & 108451 & 131035 & 20.824\% & 126654 & 16.785\% & 116840 & 7.735\% & 117784 & 8.606\% & 117440 & 8.289\% & 116938 & 7.826\% & \gray{115974} & \gray{6.937\%} \\
				X-n613-k62 & 59535 & 77555 & 30.268\% & 73633 & 23.680\% & 67545 & 13.454\% & 69069 & 16.014\% & 67296 & 13.035\% & 67730 & 13.765\% & \gray{65071} & \gray{9.299\%} \\
				X-n627-k43 & 62164 & 76776 & 23.506\% & 70744 & 13.802\% & 67523 & 8.621\% & 69361 & 11.577\% & 69915 & 12.468\% & 67896 & 9.221\% & \gray{66713} & \gray{7.318\%} \\
				X-n641-k35 & 63684 & 83138 & 30.548\% & 71986 & 13.036\% & 70631 & 10.909\% & 73624 & 15.608\% & 72517 & 13.870\% & 71974 & 13.017\% & \gray{68838} & \gray{8.093\%} \\
				X-n655-k131 & 106780 & 120771 & 13.103\% & 118758 & 11.217\% & 112289 & 5.159\% & 110657 & 3.631\% & 112598 & 5.449\% & \gray{110267} & \gray{3.266\%} & 110738 & 3.707\% \\
				X-n670-k130 & 146332 & 183183 & 25.183\% & 168210 & 14.951\% & 168829 & 15.374\% & 161571 & 10.414\% & \gray{160473} & \gray{9.664\%} & 163051 & 11.425\% & 161974 & 10.689\% \\
				X-n685-k75 & 68205 & 92701 & 35.915\% & 82607 & 21.116\% & 77890 & 14.200\% & 78473 & 15.055\% & 76817 & 12.626\% & 77132 & 13.088\% & \gray{73958} & \gray{8.435\%} \\
				X-n701-k44 & 81923 & 92723 & 13.183\% & 89704 & 9.498\% & 90580 & 10.567\% & 92198 & 12.542\% & 91600 & 11.812\% & 90703 & 10.717\% & \gray{87957} & \gray{7.365\%} \\
				X-n716-k35 & 43373 & 59383 & 36.912\% & 52170 & 20.282\% & 49480 & 14.080\% & 50605 & 16.674\% & 50073 & 15.446\% & 49405 & 13.907\% & \gray{47615} & \gray{9.780\%} \\
				X-n733-k159 & 136187 & 175848 & 29.122\% & 156268 & 14.745\% & 148581 & 9.101\% & 146080 & 7.264\% & 146443 & 7.531\% & 147334 & 8.185\% & \gray{145563} & \gray{6.885\%} \\
				X-n749-k98 & 77269 & 102208 & 32.276\% & 92403 & 19.586\% & 85046 & 10.065\% & 85325 & 10.426\% & 85320 & 10.419\% & 84712 & 9.633\% & \gray{83456} & \gray{8.007\%} \\
				X-n766-k71 & 114417 & 132968 & 16.213\% & 130101 & 13.708\% & 129866 & 13.502\% & 127752 & 11.655\% & 126515 & 10.573\% & 126387 & 10.462\% & \gray{125088} & \gray{9.326\%} \\
				X-n783-k48 & 72386 & 108577 & 49.997\% & 96432 & 33.219\% & 82839 & 14.441\% & 87562 & 20.965\% & 87027 & 20.226\% & 83864 & 15.857\% & \gray{80040} & \gray{10.574\%} \\
				X-n801-k40 & 73311 & 92125 & 25.663\% & 87187 & 18.928\% & 86121 & 17.474\% & 94076 & 28.325\% & 90039 & 22.818\% & 89478 & 22.053\% & \gray{80664} & \gray{10.030\%} \\
				X-n819-k171 & 158121 & 192102 & 21.491\% & 178856 & 13.113\% & 174446 & 10.324\% & 172387 & 9.022\% & 172578 & 9.143\% & 171676 & 8.573\% & \gray{169066} & \gray{6.922\%} \\
				X-n837-k142 & 193737 & 231002 & 19.235\% & 230226 & 18.834\% & 208669 & 7.707\% & 209540 & 8.157\% & 208974 & 7.865\% & 209031 & 7.894\% & \gray{205696} & \gray{6.173\%} \\
				X-n856-k95 & 88965 & 117243 & 31.786\% & 105763 & 18.882\% & 98164 & 10.340\% & 102312 & 15.003\% & 100818 & 13.324\% & 98914 & 11.183\% & \gray{96218} & \gray{8.153\%} \\
				X-n876-k59 & 99299 & 114212 & 15.018\% & 114175 & 14.981\% & 107477 & 8.236\% & 109693 & 10.467\% & 108742 & 9.509\% & 110843 & 11.625\% & \gray{105502} & \gray{6.247\%} \\
				X-n895-k37 & 53860 & 106062 & 96.922\% & 70363 & 30.641\% & 64225 & 19.244\% & 73280 & 36.056\% & 70673 & 31.216\% & 67830 & 25.938\% & \gray{61290} & \gray{13.795\%} \\
				X-n916-k207 & 329179 & 387367 & 17.677\% & 374899 & 13.889\% & 353039 & 7.248\% & 351887 & 6.898\% & 358336 & 8.857\% & 352488 & 7.081\% & \gray{347539} & \gray{5.578\%} \\
				X-n936-k151 & 132715 & 200816 & 51.314\% & 161700 & 21.840\% & 162903 & 22.746\% & 154847 & 16.676\% & 153747 & 15.847\% & 155618 & 17.257\% & \gray{152952} & \gray{15.248\%} \\
				X-n957-k87 & 85465 & 126220 & 47.686\% & 124190 & 45.311\% & 103089 & 20.621\% & 108664 & 27.144\% & 105588 & 23.522\% & 106903 & 25.084\% & \gray{93710} & \gray{9.647\%} \\
				X-n979-k58 & 118976 & 138987 & 16.819\% & 132651 & 11.494\% & 129633 & 8.957\% & 133201 & 11.956\% & 137728 & 15.761\% & 132728 & 11.559\% & \gray{127096} & \gray{6.825\%} \\
				X-n1001-k43 & 72355 & 132976 & 83.783\% & 89175 & 23.246\% & 85852 & 18.654\% & 92974 & 28.497\% & 90653 & 25.287\% & 93476 & 29.191\% & \gray{81075} & \gray{12.052\%} \\
				\midrule
				\multicolumn{2}{c|}{Avg. Gap ($501 < n \leq 1001$)} & \multicolumn{2}{c}{30.190\%} & \multicolumn{2}{c}{18.918\%} & \multicolumn{2}{c}{12.252\%} & \multicolumn{2}{c}{14.199\%} & \multicolumn{2}{c}{13.135\%} & \multicolumn{2}{c}{12.814\%} & \multicolumn{2}{c}{\gray{8.215\%}} \\
				\midrule
				\multicolumn{2}{c|}{Avg. Gap} & \multicolumn{2}{c}{15.863\%} & \multicolumn{2}{c}{11.693\%} & \multicolumn{2}{c}{8.428\%} & \multicolumn{2}{c}{9.229\%} & \multicolumn{2}{c}{8.586\%} & \multicolumn{2}{c}{8.441\%} & \multicolumn{2}{c}{\gray{5.860\%}} \\
				\bottomrule
		\end{tabular}}
	\end{table}

\section{Results on public benchmarks}
\label{app:public}
	
	To assess cross-size and cross-distribution generalization, we evaluate all neural solvers on the Set-X instances of CVRPLib \citep{Uchoa2017}. These CVRP instances differ significantly from training data, in terms of node distribution, demand patterns and graph scales that range well beyond the training size. Following prior MTL practice \citep{Correa2026filmmed,Pan2025decomposable}, we use the models trained on $n{=}100$ without any fine-tuning on Set-X.
	
	Table~\ref{tab:setX_results} reports the objective values and gaps to the best-known solutions in the literature. Averaged over the full Set-X dataset, our method attains a mean gap of $5.860\%$, which improves upon all evaluated state-of-the-art neural baselines ($30.5\%$ and $30.6\%$ relative average-gap reductions compared to RouteFinder and MoSES(CaDA), respectively). Instance-wise, our method achieves the lowest gap among neural solvers on 81 of the 100 instances. These improvements hold across the three reported problem-size ranges. On instances with fewer than 251 customers the mean gap is $3.710\%$, and it remains the lowest among neural solvers for mid-scale ($5.680\%$) and large-scale instances ($8.215\%$). The latter result is particularly informative. Even though all models were trained only at $n{=}100$, the proposed approach degrades less as graph size grows, indicating stronger cross-size generalization than prior MTL solvers.

		\begin{table}[H]
		\centering
		\caption{Zero-shot generalization performance on 32 unseen VRP variants. Gray-shaded cells indicate the best objective and gap among all neural models.}
		\label{tab:zeroshot}
		\resizebox{\linewidth}{!}{% 
			\large
			\renewcommand{\arraystretch}{1.05}
			\begin{tabular}{l|cc|cc|cc|cc|cc|cc|cc|cc|cc}
				\toprule
				& \multicolumn{2}{c|}{PyVRP} & \multicolumn{2}{c|}{OR-Tools} & \multicolumn{2}{c|}{MTPOMO} & \multicolumn{2}{c|}{MVMoE} & \multicolumn{2}{c|}{RouteFinder} & \multicolumn{2}{c|}{CaDA} & \multicolumn{2}{c|}{FiLMMeD(CaDA)} & \multicolumn{2}{c|}{MoSES(CaDA)} & \multicolumn{2}{c}{Ours} \\
				\cmidrule(lr){2-3} \cmidrule(lr){4-5} \cmidrule(lr){6-7} \cmidrule(lr){8-9} \cmidrule(lr){10-11} \cmidrule(lr){12-13} \cmidrule(lr){14-15} \cmidrule(lr){16-17} \cmidrule(lr){18-19}
				Variant & Obj. & Gap & Obj. & Gap & Obj. & Gap & Obj. & Gap & Obj. & Gap & Obj. & Gap & Obj. & Gap & Obj. & Gap & Obj. & Gap \\
				\midrule
				VRPMB & 13.54 & * & 14.93 & 10.27\% & 15.04 & 11.32\% & 14.99 & 10.94\% & 14.88 & 10.13\% & 14.83 & 9.73\% & 14.65 & 8.37\% & 14.93 & 10.50\% & \cellcolor{gray!25}14.63 & \cellcolor{gray!25}8.23\% \\
				OVRPMB & 9.01 & * & 10.59 & 17.54\% & 10.87 & 20.65\% & 10.85 & 20.42\% & 10.72 & 19.02\% & 10.58 & 17.38\% & 10.51 & 16.71\% & 10.73 & 19.09\% & \cellcolor{gray!25}10.41 & \cellcolor{gray!25}15.53\% \\
				VRPMBL & 13.78 & * & 15.42 & 11.90\% & 15.41 & 11.97\% & 15.33 & 11.37\% & 15.18 & 10.32\% & 15.05 & 9.27\% & 14.97 & 8.71\% & 15.23 & 10.65\% & \cellcolor{gray!25}14.94 & \cellcolor{gray!25}8.43\% \\
				VRPMBTW & 25.51 & * & 29.97 & 17.48\% & 28.31 & 11.06\% & 28.32 & 11.10\% & 28.29 & 10.87\% & 28.40 & 11.42\% & 28.28 & 10.92\% & 28.40 & 11.40\% & \cellcolor{gray!25}28.22 & \cellcolor{gray!25}10.70\% \\
				OVRPMBL & 9.01 & * & 10.59 & 17.54\% & 10.85 & 20.43\% & 10.82 & 20.14\% & 10.72 & 19.01\% & 10.58 & 17.41\% & 10.51 & 16.72\% & 10.72 & 18.96\% & \cellcolor{gray!25}10.41 & \cellcolor{gray!25}15.54\% \\
				OVRPMBTW & 16.97 & * & 19.31 & 13.79\% & 18.51 & 9.08\% & 18.55 & 9.33\% & 18.45 & 8.68\% & 18.57 & 9.44\% & 18.56 & 9.38\% & 18.48 & 8.91\% & \cellcolor{gray!25}18.43 & \cellcolor{gray!25}8.63\% \\
				VRPMBLTW & 25.85 & * & 30.44 & 17.76\% & 28.73 & 11.27\% & 28.70 & 11.16\% & 28.65 & 10.82\% & 28.75 & 11.38\% & 28.62 & 10.86\% & 28.69 & 11.12\% & \cellcolor{gray!25}28.59 & \cellcolor{gray!25}10.76\% \\
				OVRPMBLTW & 16.97 & * & 19.31 & 13.79\% & 18.51 & 9.12\% & 18.55 & 9.30\% & 18.45 & 8.69\% & 18.57 & 9.43\% & 18.55 & 9.32\% & 18.47 & 8.82\% & \cellcolor{gray!25}18.43 & \cellcolor{gray!25}8.63\% \\
				MDVRP & 11.89 & * & 12.52 & 5.30\% & 16.07 & 35.74\% & 16.02 & 35.35\% & 15.98 & 35.02\% & 17.16 & 45.33\% & 14.30 & 20.69\% & 18.26 & 55.01\% & \cellcolor{gray!25}13.97 & \cellcolor{gray!25}17.80\% \\
				MDOVRP & 7.97 & * & 8.16 & 2.38\% & 10.28 & 29.06\% & 10.24 & 28.59\% & 10.18 & 27.82\% & \cellcolor{gray!25}9.74 & \cellcolor{gray!25}22.31\% & 9.84 & 23.70\% & 10.20 & 28.14\% & 9.97 & 25.06\% \\
				MDVRPB & 11.64 & * & 12.22 & 4.98\% & 15.18 & 30.66\% & 15.12 & 30.13\% & 15.05 & 29.53\% & 17.86 & 54.31\% & 14.44 & 24.38\% & 17.38 & 50.11\% & \cellcolor{gray!25}14.27 & \cellcolor{gray!25}22.74\% \\
				MDVRPL & 11.90 & * & 12.52 & 5.21\% & 16.30 & 37.58\% & 16.25 & 37.17\% & 16.20 & 36.76\% & 17.14 & 45.04\% & 14.37 & 21.27\% & 18.31 & 55.35\% & \cellcolor{gray!25}14.00 & \cellcolor{gray!25}17.99\% \\
				MDVRPTW & 19.33 & * & 19.62 & 1.50\% & 26.68 & 38.56\% & 26.67 & 38.51\% & 26.51 & 37.64\% & 27.18 & 41.22\% & 24.12 & 25.25\% & 32.21 & 67.90\% & \cellcolor{gray!25}23.41 & \cellcolor{gray!25}21.48\% \\
				MDOVRPTW & 13.00 & * & 13.09 & 0.69\% & 17.57 & 35.67\% & 17.57 & 35.68\% & 17.48 & 34.96\% & 17.23 & 33.02\% & 15.93 & 23.03\% & 18.48 & 42.84\% & \cellcolor{gray!25}15.59 & \cellcolor{gray!25}20.28\% \\
				MDOVRPB & 8.69 & * & 8.87 & 2.07\% & 10.94 & 26.08\% & 10.89 & 25.56\% & 10.82 & 24.74\% & 10.93 & 26.06\% & \cellcolor{gray!25}10.53 & \cellcolor{gray!25}21.43\% & 11.34 & 30.81\% & 10.77 & 24.04\% \\
				MDOVRPL & 7.97 & * & 8.16 & 2.38\% & 10.28 & 29.07\% & 10.24 & 28.60\% & 10.18 & 27.84\% & \cellcolor{gray!25}9.74 & \cellcolor{gray!25}22.26\% & 9.93 & 24.73\% & 10.20 & 28.17\% & 9.88 & 24.01\% \\
				MDVRPBL & 11.68 & * & 12.22 & 4.62\% & 15.80 & 35.54\% & 15.73 & 34.95\% & 15.62 & 33.98\% & 17.50 & 50.52\% & 14.47 & 24.15\% & 17.79 & 53.04\% & \cellcolor{gray!25}14.31 & \cellcolor{gray!25}22.65\% \\
				MDVRPBTW & 22.03 & * & 22.40 & 1.68\% & 30.55 & 39.23\% & 30.55 & 39.22\% & 30.36 & 38.36\% & 30.88 & 40.84\% & 27.54 & 25.59\% & 36.02 & 64.66\% & \cellcolor{gray!25}26.43 & \cellcolor{gray!25}20.35\% \\
				MDVRPLTW & 19.35 & * & 19.66 & 1.60\% & 27.13 & 40.71\% & 27.12 & 40.67\% & 26.93 & 39.69\% & 27.63 & 43.44\% & 24.24 & 25.73\% & 31.90 & 66.02\% & \cellcolor{gray!25}23.49 & \cellcolor{gray!25}21.72\% \\
				MDOVRPBL & 8.69 & * & 8.87 & 2.07\% & 10.94 & 26.11\% & 10.90 & 25.66\% & 10.82 & 24.78\% & 10.93 & 25.99\% & \cellcolor{gray!25}10.65 & \cellcolor{gray!25}22.85\% & 11.31 & 30.50\% & 10.68 & 23.06\% \\
				MDOVRPBTW & 14.37 & * & 14.49 & 0.84\% & 19.69 & 37.62\% & 19.69 & 37.62\% & 19.59 & 36.95\% & 19.38 & 35.44\% & 17.75 & 24.09\% & 21.25 & 48.58\% & \cellcolor{gray!25}17.24 & \cellcolor{gray!25}20.35\% \\
				MDOVRPLTW & 13.00 & * & 13.09 & 0.69\% & 17.58 & 35.70\% & 17.58 & 35.74\% & 17.48 & 34.95\% & 17.22 & 32.99\% & 15.93 & 22.98\% & 18.42 & 42.37\% & \cellcolor{gray!25}15.55 & \cellcolor{gray!25}19.97\% \\
				MDVRPBLTW & 22.06 & * & 22.43 & 1.68\% & 31.09 & 41.52\% & 31.06 & 41.39\% & 30.86 & 40.49\% & 31.57 & 43.84\% & 27.62 & 25.79\% & 35.82 & 63.58\% & \cellcolor{gray!25}26.53 & \cellcolor{gray!25}20.62\% \\
				MDOVRPBLTW & 14.37 & * & 14.49 & 0.84\% & 19.69 & 37.64\% & 19.69 & 37.61\% & 19.60 & 36.96\% & 19.38 & 35.44\% & 17.78 & 24.23\% & 21.23 & 48.44\% & \cellcolor{gray!25}17.21 & \cellcolor{gray!25}20.11\% \\
				MDVRPMB & 10.68 & * & 12.22 & 14.42\% & 15.14 & 42.22\% & 15.08 & 41.67\% & 14.99 & 40.80\% & 17.51 & 65.10\% & 14.24 & 33.86\% & 17.34 & 63.55\% & \cellcolor{gray!25}14.14 & \cellcolor{gray!25}32.80\% \\
				MDOVRPMB & 7.66 & * & 8.88 & 15.93\% & 10.91 & 42.57\% & 10.90 & 42.41\% & 10.77 & 40.67\% & 10.78 & 40.93\% & \cellcolor{gray!25}10.42 & \cellcolor{gray!25}36.20\% & 11.18 & 46.12\% & 10.52 & 37.40\% \\
				MDVRPMBL & 10.71 & * & 12.23 & 14.19\% & 15.49 & 45.23\% & 15.40 & 44.37\% & 15.28 & 43.27\% & 16.86 & 58.45\% & 14.17 & 32.82\% & 17.41 & 63.80\% & \cellcolor{gray!25}14.04 & \cellcolor{gray!25}31.52\% \\
				MDVRPMBTW & 19.29 & * & 22.39 & 16.07\% & 28.44 & 48.01\% & 28.46 & 48.12\% & 28.43 & 47.93\% & 29.38 & 52.98\% & 26.24 & 36.64\% & 31.74 & 65.65\% & \cellcolor{gray!25}24.97 & \cellcolor{gray!25}29.82\% \\
				MDOVRPMBL & 7.66 & * & 8.87 & 15.80\% & 10.90 & 42.45\% & 10.88 & 42.13\% & 10.76 & 40.62\% & 10.78 & 40.91\% & 10.52 & 37.47\% & 11.16 & 45.94\% & \cellcolor{gray!25}10.46 & \cellcolor{gray!25}36.57\% \\
				MDOVRPMBTW & 12.96 & * & 14.49 & 11.81\% & 18.56 & 43.63\% & 18.61 & 44.04\% & 18.49 & 43.14\% & 18.70 & 44.76\% & 17.19 & 33.09\% & 19.24 & 49.02\% & \cellcolor{gray!25}16.50 & \cellcolor{gray!25}27.57\% \\
				MDVRPMBLTW & 19.31 & * & 22.43 & 16.16\% & 28.93 & 50.36\% & 28.89 & 50.19\% & 28.80 & 49.69\% & 29.95 & 55.78\% & 26.26 & 36.57\% & 31.64 & 64.92\% & \cellcolor{gray!25}25.01 & \cellcolor{gray!25}29.88\% \\
				MDOVRPMBLTW & 12.96 & * & 14.49 & 11.81\% & 18.56 & 43.65\% & 18.60 & 43.95\% & 18.50 & 43.17\% & 18.70 & 44.78\% & 17.20 & 33.16\% & 19.20 & 48.71\% & \cellcolor{gray!25}16.49 & \cellcolor{gray!25}27.50\% \\
				\midrule
				Avg. Gap & \multicolumn{2}{c|}{*} & \multicolumn{2}{c|}{8.59\%} & \multicolumn{2}{c|}{31.86\%} & \multicolumn{2}{c|}{31.66\%} & \multicolumn{2}{c|}{30.85\%} & \multicolumn{2}{c|}{34.29\%} & \multicolumn{2}{c|}{23.46\%} & \multicolumn{2}{c|}{41.33\%} & \multicolumn{2}{c|}{\cellcolor{gray!25}21.30\%} \\
				\bottomrule
			\end{tabular}
		}
	\end{table}

\section{Zero-shot evaluation on unseen variants}
	\label{app:zeroshot}
	
	To test how well the model transfers to constraint combinations not seen during training, we evaluate all methods on 32 additional variants that extend the 16 training variants with mixed backhauls (MB), multi-depots (MD), or both. The encoder already encodes MB and MD through the constraint flag vector $z$, but these attributes are never activated during training. 	Models are therefore tested in a strict zero-shot setting, without fine-tuning, using the same greedy POMO decoding protocol as in the main experiments.
	
	Table~\ref{tab:zeroshot} reports objective values and gaps to PyVRP for every unseen variant at $n{=}100$. Our method achieves the best neural objective and gap on 27 of the 32 variants, outperforming the strongest prior neural baseline (FiLMMeD(CaDA)). It also outperforms every other neural competitor on at least 28 variants, including 32 over MTPOMO, MVMoE, and RouteFinder, 30 over CaDA, and 32 over MoSES(CaDA). On the eight mixed-backhaul variants, our model ranks first throughout. On multi-depot instances, it achieves the lowest neural gap on 19 of the 24 MD variants, while FiLMMeD(CaDA) or CaDA lead on the remaining five cases.

		\begin{figure}[H]
		\centering
		\includegraphics[width=\linewidth]{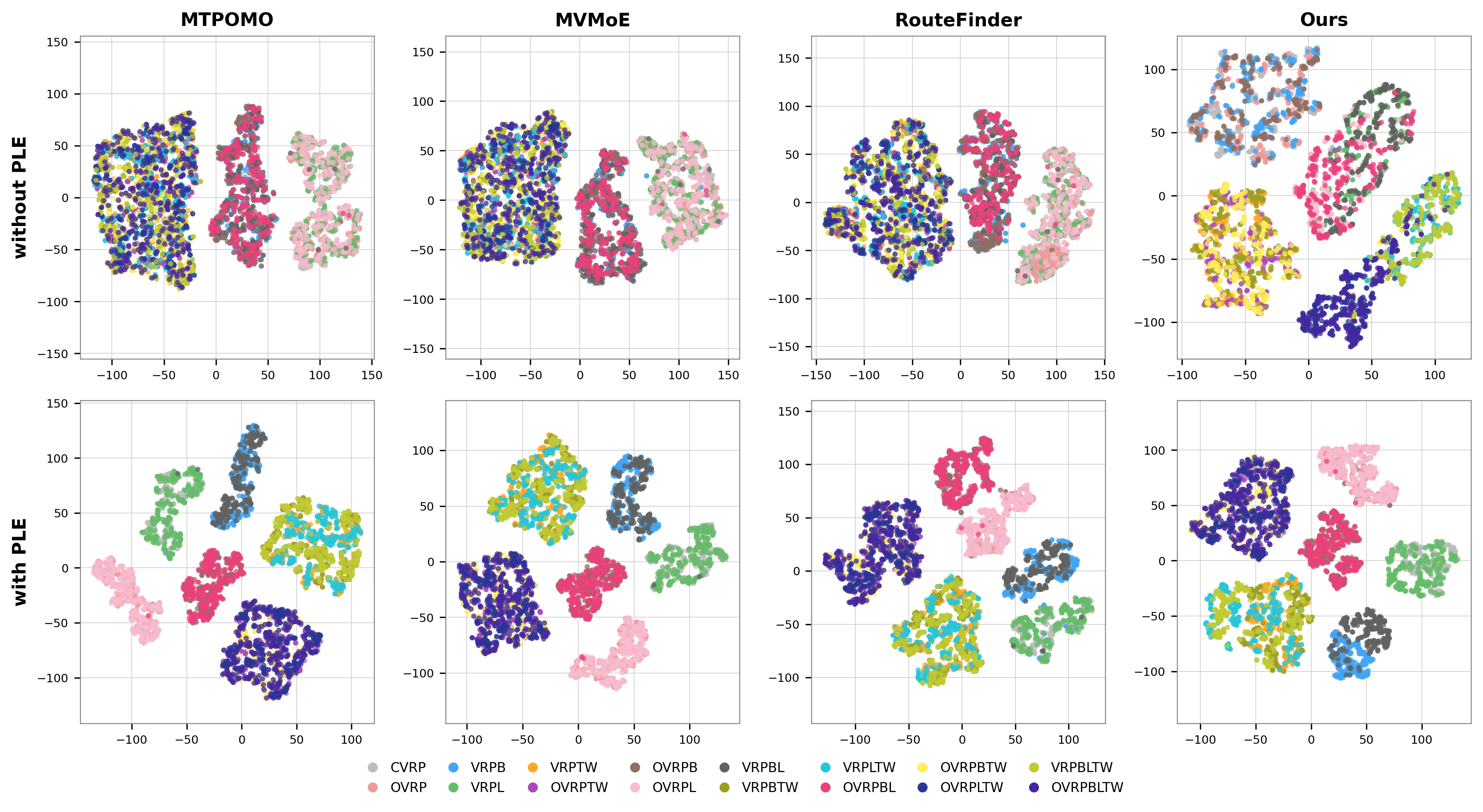}
		\caption{t-SNE of final encoder embeddings on the 16 training variants ($n{=}50$). Columns compare MTPOMO, MVMoE, RouteFinder, and our model. The top row shows shared-encoder baselines; the bottom row shows the same backbones with PLE. Points are colored by variant.}
		\label{fig:representation_tsne}
	\end{figure}

\section{Representation geometry}
	\label{app:representation_geometry}
	
	In Section~\ref{sec:model_architecture}, we motivate PLE as a structured alternative to encoding heterogeneous VRP constraints through one fully shared stack. Here, we provide a complementary geometric view of the learned encoder representations on the 16 main in-distribution training variants. We apply t-SNE \citep{Maaten2008} to the final node embeddings of the encoder for MTPOMO, MVMoE, RouteFinder, and our model, each trained at $n{=}50$ with and without PLE. 	Figure~\ref{fig:representation_tsne} shows the results, with different colors points per variant.
	
	Without PLE, embeddings from all four backbones mix across variants and form a few broad overlapping clusters. With PLE, the same backbones show much more separated clusters that are tighter and more homogeneous. This contrast indicates that PLE succeeds in reducing representation entanglement in the encoded embeddings used by the decoder.

\end{document}